\pdfoutput=1

\documentclass[11pt]{article}

\usepackage[final]{acl}
\usepackage{times}
\usepackage{latexsym}

\usepackage[utf8]{inputenc} 
\usepackage[T1]{fontenc}    
\usepackage{hyperref}       
\usepackage{url}            
\usepackage{booktabs}       
\usepackage{amsfonts}       
\usepackage{nicefrac}       
\usepackage{microtype}      
\usepackage{times}
\usepackage{latexsym}

\usepackage[T1]{fontenc}

\usepackage[utf8]{inputenc}
\usepackage{inconsolata}
\usepackage{microtype}

\usepackage{inconsolata}
\usepackage{hyperref}
\usepackage{url}
\usepackage{hyperref}       
\usepackage{url}            
\usepackage{booktabs}       
\usepackage{amsfonts}       
\usepackage{nicefrac}       
\usepackage{microtype}      
\usepackage{url}            
\usepackage{booktabs}       
\usepackage{amsfonts}       
\usepackage{nicefrac}       
\usepackage{microtype}      
\usepackage{subcaption}
\usepackage{amsmath,amsfonts,amssymb}
\usepackage{breqn}
\usepackage{tabularx}
\usepackage{multirow}
\usepackage{graphicx}
\usepackage{color}
\usepackage{tcolorbox}
\usepackage{CJK}
\usepackage{adjustbox}
\usepackage{colortbl}
\usepackage{multicol}
\usepackage{vwcol} 
\newsavebox{\algleft}
\newsavebox{\algright}
\usepackage{bm}
\usepackage{booktabs}
\usepackage{amsmath,stackengine}
\usepackage[linesnumbered,ruled,vlined]{algorithm2e}
\usepackage{times}
\usepackage{latexsym}
\usepackage{bbm}
\usepackage[T1]{fontenc}    
\usepackage{hyperref}       
\usepackage{url}            
\usepackage{booktabs}       
\usepackage{amsfonts}       
\usepackage{nicefrac}       
\usepackage{microtype}      
\usepackage{amssymb}
\usepackage{pifont}
\usepackage{bbm}
\usepackage{breqn}
\usepackage{longtable}
\usepackage{algpseudocode}
\usepackage{enumitem}
\usepackage{microtype}
\usepackage{enumitem}
\usepackage{amsmath}
\usepackage{chngcntr} 
\usepackage{wrapfig}
\usepackage{listings}

\definecolor{jsonkey}{rgb}{0.6, 0.2, 0.2}
\definecolor{jsonstring}{rgb}{0.2, 0.6, 0.2}

\lstdefinestyle{jsonstyle}{
    basicstyle=\ttfamily\small,
    stringstyle=\color{jsonstring},
    keywordstyle=\color{jsonkey},
    showstringspaces=false,
    breaklines=true,
    frame=none,
    morekeywords={true,false,null} 
}

\definecolor{BlueBG}{rgb}{0,0.46,0.71}
\newcommand{\ourmodel}{{Llama-Fin}\xspace} 
\newcommand{\ourframework}{{\sc{FinDaP}}\xspace} 

\title{Demystifying Domain-adaptive Post-training for Financial LLMs}

\author{Zixuan Ke, Yifei Ming, Xuan-Phi Nguyen, Caiming Xiong, Shafiq Joty \\ 
Salesforce AI Research\\
\texttt{\{zixuan.ke,cxiong,sjoty\}@salesforce.com}\\
\includegraphics[width=0.4cm]{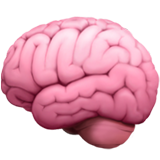} \small Project Page: \url{https://vincent950129.github.io/adapt-llm/} \\
\includegraphics[width=0.4cm]{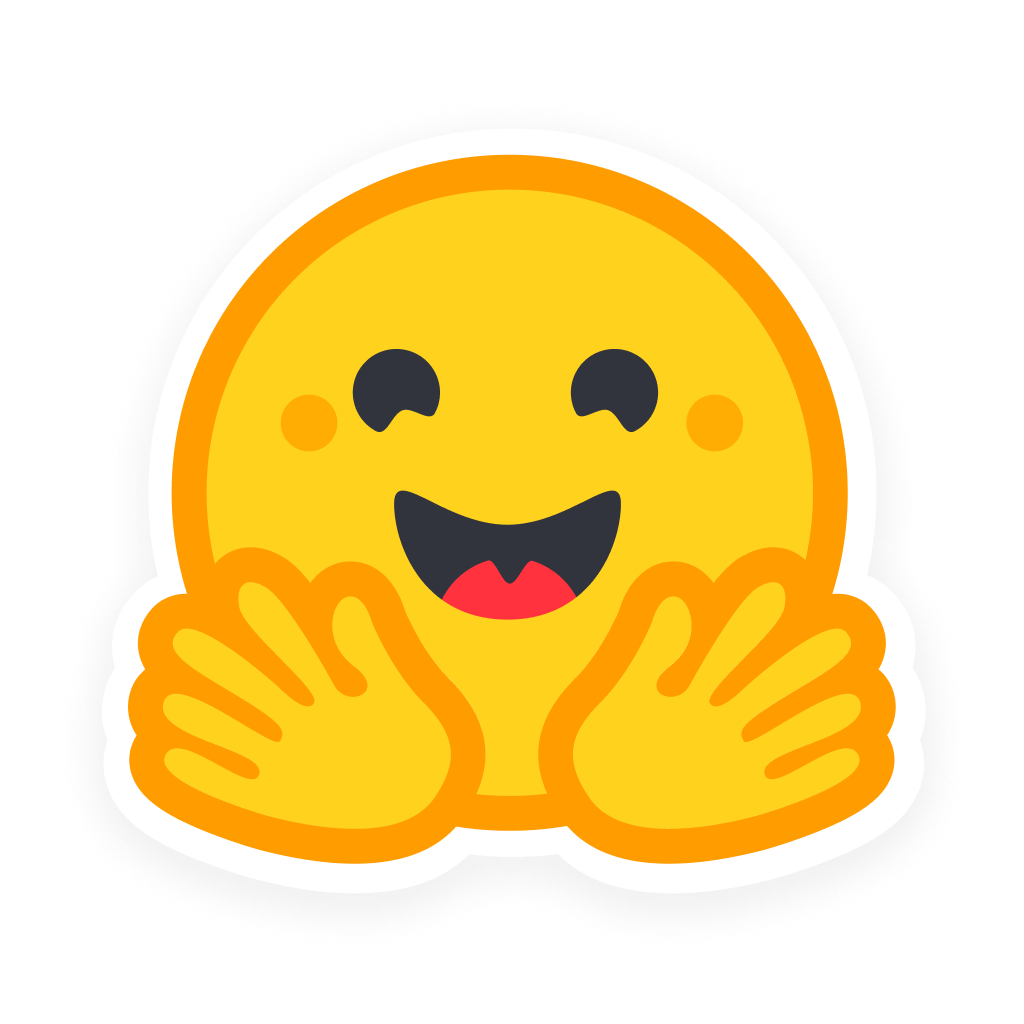} \small \ourmodel-8b: \url{https://huggingface.co/Salesforce/Llama-Fin-8b} \\
\includegraphics[width=0.4cm]{emoji/hf.png} \small FinTrain: \url{https://huggingface.co/datasets/Salesforce/FinTrain} \\
\includegraphics[width=0.4cm]{emoji/hf.png} \small FinEval: \url{https://huggingface.co/datasets/Salesforce/FinEval} \\
\includegraphics[width=0.4cm]{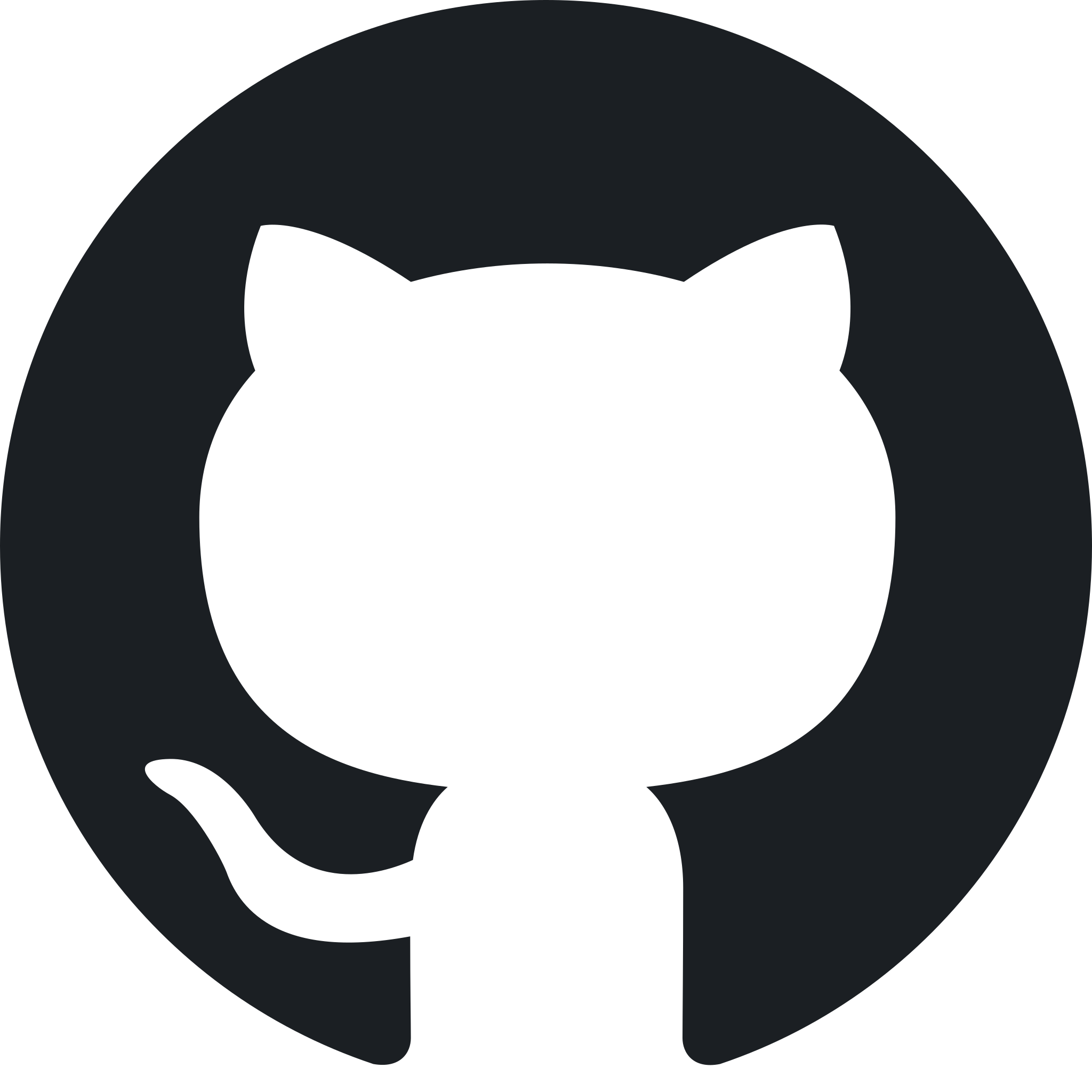} \small Code: \url{https://github.com/SalesforceAIResearch/FinDAP} 
}

\definecolor{innerboxcolor}{HTML}{00B2E6} 
\definecolor{outerboxcolor}{HTML}{D7F6FF} 
\definecolor{lightblue}{HTML}{D7F6FF}

\usepackage{tikz}
\usetikzlibrary{decorations.pathmorphing}

\begin{document}
\maketitle

\begin{abstract}

Domain-adaptive post-training of large language models (LLMs) has emerged as a promising approach for specialized domains such as medicine and finance. However, significant challenges remain in identifying optimal adaptation criteria and training strategies across varying data and model configurations.
To address these challenges, we introduce \ourframework, a systematic and fine-grained investigation into domain-adaptive post-training of LLMs for the finance domain. {Our approach consists of four key components: \textit{FinCap}, which defines the core capabilities required for the target domain; \textit{FinRec}, an effective training recipe that jointly optimizes continual pre-training and instruction-following, along with a novel preference data distillation method leveraging process signals from a generative reward model; \textit{FinTrain}, a curated set of training datasets supporting FinRec; and \textit{FinEval}, a comprehensive evaluation suite aligned with FinCap.}
The resulting model, Llama-Fin, achieves state-of-the-art performance across a wide range of financial tasks. Our analysis also highlights how each post-training stage contributes to distinct capabilities, uncovering specific challenges and effective solutions, providing valuable insights for domain adaptation of LLMs.


\end{abstract}

\section{Introduction}

While LLMs have demonstrated strong generalization across a variety of tasks, they often struggle to perform well in specialized domains such as finance and law. 
Consequently, \emph{domain-adaptive post-training} of LLMs has garnered significant attention recently \citep{colombo2024saullmb, xie-etal-2024-efficient}. In the earlier days of language models, \emph{continual pre-training} (CPT) was the dominant strategy. This involved further training a pre-trained model on domain-specific plain text and then fine-tuning it for individual tasks \citep{gururangan-etal-2020-dont, ke2023dgs}. With LLMs, the post-training focus has shifted to zero- and few-shot task generalization through  methods such as  \textit{instruction-tunning (IT)} (aka. supervised fine-tuning or SFT) and \textit{preference alignment ({PA})}. 
While prompt engineering of powerful general LLMs with zero- or few-shot examples has emerged as a convenient approach to adapting them to new tasks, to get the most optimal performance on a target domain, recent methods explore  
fine-tuning model wights to make them domain experts \cite{chen2023meditron,lwdc23,colombo2024saullm}.



Building on this trend, this work focuses on adapting LLMs to specific domains \textit{through parameter training}. It complements semi-parametric methods that leverage \textit{external} knowledge, such as retrieval-augmented generation \citep{lewis2020retrieval,bridging_retriever_llm_ke2024}. 
Our focus is also different from general post-training, as the goal is not to develop another general-purpose  LLM but to create specialized, expert-level LLMs tailored to a specific domain. By focusing on a specific domain, we develop models that are not only more compact in size but also deliver significantly more accurate and contextually relevant responses compared to general-purpose LLMs. Their smaller size enhances efficiency, optimizing both computational resource usage and training time.


Despite the potential of domain-specific LLMs, there is still no systematic study on what makes a good domain-specific LLM. In this work, we consider \emph{finance} as the domain of interest and aim to address the following research questions:

\vspace{-2mm}
\begin{tcolorbox}[colback=outerboxcolor,colframe=innerboxcolor,fonttitle=\bfseries,arc=0.5mm,boxrule=0.2pt,boxrule=-1pt]
Given a strong general-purpose LLM (\emph{e.g.}, Llama3-8b-inst), how to {effectively adapt it to a target domain} (\emph{e.g.}, finance) by post-training? What criteria are desirable for successful adaptation? What are effective training recipes with respect to data and model? 
\end{tcolorbox}
\vspace{-2mm}

\begin{figure*}[t]
\centering
\includegraphics[width=0.85\textwidth]{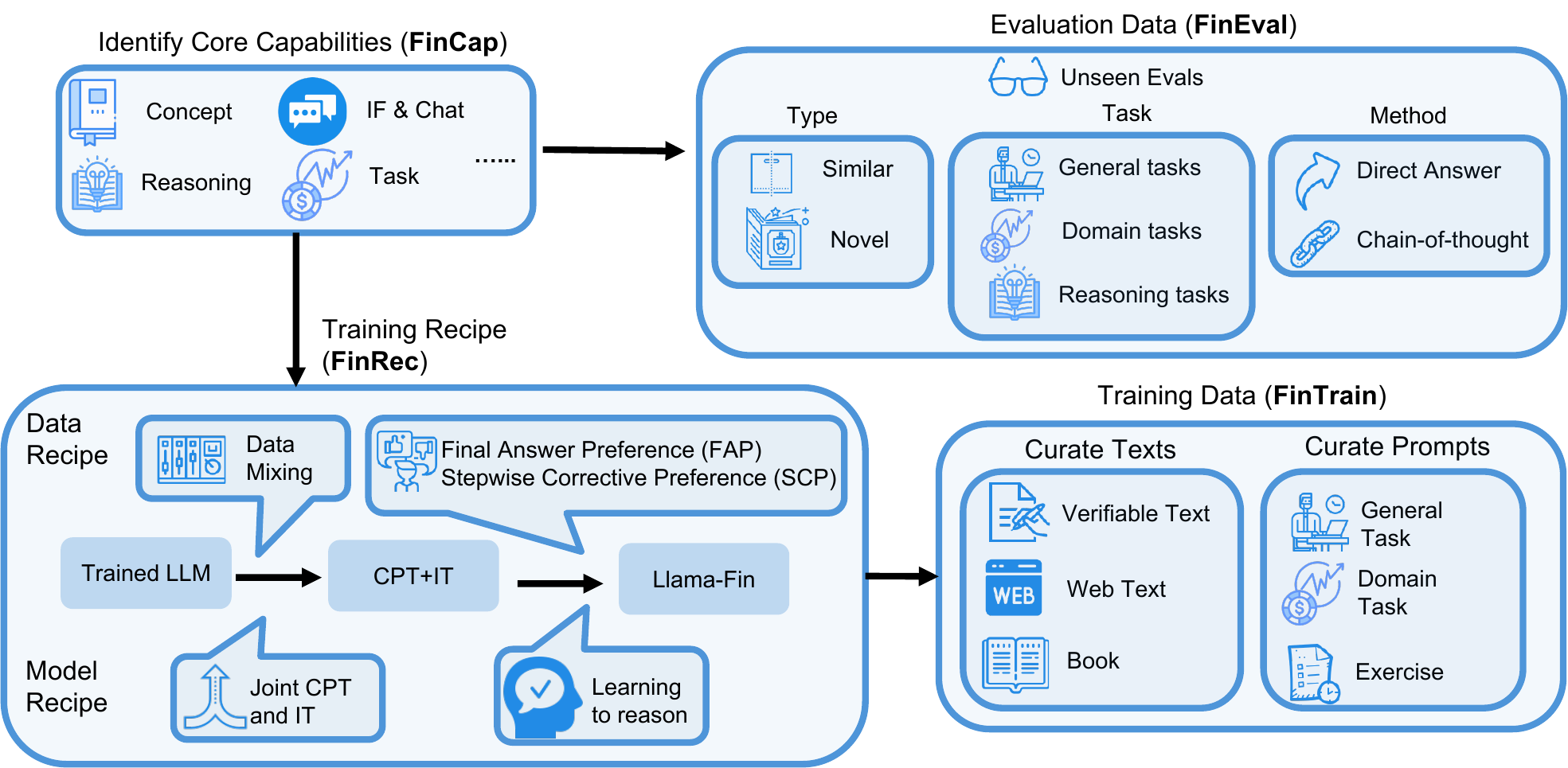}
\vspace{-3mm}
\caption{
{An overview of our finance-specific post-training framework, \ourframework.   It comprises four key components: (1) \textbf{FinCap}, the core expected capabilities, including concepts, reasoning, instruction-following and tasks; (2) \textbf{FinRec}, encompassing both data and model strategies to guide domain-adaptive post-training; (3) \textbf{FinTrain}, which curates training texts and prompts based on the data recipe; and (4) \textbf{FinEval}, a comprehensive evaluation framework designed to assess performance on unseen tasks, categorized into similar and novel, general and domain-specific, and reasoning tasks, using both direct-answer and chain-of-thought (CoT) evaluation methods.
}
}
\label{fig.overview}
\vspace{-5mm}
\end{figure*}


Prior studies~\cite{bhatia2024fintralfamilygpt4level,xie2024openfinllmsopenmultimodallarge} typically adopt a simplified and informal framework (see \S\ref{sec.related_work}) in that they evaluate only on a set of domain-specific end tasks such as {sentiment analysis and NER}, and {they simply follow standard post-training stages (CPT, IT and/or PA) without considering their impact or optimizing their recipe for domain-adaptive post-training.} 
This simplified approach can misalign with our broader expectations for a domain-expert LLM. A domain-expert LLM should not only excel at such end tasks but also achieve broader capabilities, such as follow task instructions effectively and reason in a way that aligns with domain-specific knowledge, while retaining general capabilities. 


We argue that domain-adaptive post-training poses unique challenges compared to pre-training or general post-training. There are multiple factors to be considered: \textbf{(1)} For a particular target domain, it is essential to establish the \textbf{desirable capabilities} that a domain-expert LLM should possess, as these capabilities serve as a guiding framework for the entire adaptation process; \textbf{(2)} The training recipe should be tailored specifically to adapt an already trained LLM (e.g., Llama3-inst) through post-training. This differs from training a model from scratch or from a base pre-trained checkpoint, as it requires careful consideration of \textbf{catastrophic forgetting}  and \textbf{knowledge transfer} from the original LLM, which already possesses strong general knowledge and instruction following capabilities. Each of the standard CPT, IT, and PA stages have different impacts and trade-offs with respect to knowledge forgetting and transfer, {as do the in-domain, general-domain datasets, and the mixture of them.}
Moreover, it should also be designed to support the desired capabilities. For example, {improving reasoning capability might require more dense supervision than the final answer level correctness score}. \textbf{(3)} {The desired \textbf{quantity and quality of training datasets} should be carefully balanced}: high-quality general-domain data is required to mitigate forgetting, while diverse data and supervision signals are necessary to learn domain knowledge. \textbf{(4)} Finally, the \textbf{evaluation methods} should align with the desired capabilities. Different evaluation techniques may be required for certain capabilities; for example, chain-of-thought (CoT) \cite{wei2023chainofthoughtpromptingelicitsreasoning} reasoning is often necessary to effectively evaluate reasoning tasks.

{In this work, we introduce \textbf{\ourframework} (Figure \ref{fig.overview}), a novel finance-specific framework designed to incorporate all these factors in domain-specific post-training. To our knowledge, none of the prior studies consider all of them to provide a principled guidance on domain-adaptive post-training.} \ourframework integrates four key components: (1) \textbf{FinCap}, a set of core capabilities required for the domain expert LLM, {\color{black}derived from a systematic review of prior literature and input from domain experts in finance.} These include domain concepts, tasks, instruction following and reasoning; (2) \textbf{FinRec}, a training recipe that \textbf{jointly performs CPT and IT}, and subsequently conducts PA, balancing trade-offs across these stages to mitigate forgetting and improve task generalization. It also proposes to use \textbf{mixture of in-domain and general domain data} in the data recipe, alongside a novel preference alignment method for improving reasoning capability that constructs data using the preference signal in reasoning steps,  \textbf{Stepwise Corrective Preference (SCP)}, and final answer, \textbf{Final Answer Preference (FAP)}; (3) \textbf{FinTrain}, a curated set of training datasets implementing FinRec, which carefully balances quality  and diversity; and (4) \textbf{FinEval}, a comprehensive evaluation framework covering a wide range of tasks, including reasoning tasks assessed through CoT.
\footnote{We will open-source the data, checkpoint, code, leaderboard for all components upon acceptance.}


We apply \ourframework  on the instruction-tuned Llama3-8b-instruct \citep{grattafiori2024llama3herdmodels}. Our best performing recipe yields \textbf{\ourmodel} that outperforms all considered baselines, including large open models at the 70B scale and proprietary models like GPT-4o, on tasks that are similar (yet unseen) to the training data. Even on novel tasks that were never encountered in training, \ourmodel remains competitive and consistently outperforms its base model across all identified capabilities. {\color{black}In summary, our key contributions are:
\begin{itemize}
[leftmargin=*,noitemsep,topsep=5pt]
    \item \textbf{Comprehensive guidance} for finance-specific post-training, including identification of capabilities, evaluation, data and model recipe design.
    \item \textbf{Systematic exploration} on each stage of post-training, with an emphasis on the goals, challenges and effective approaches. 
    \item \textbf{Novel preference alignment approach} that constructs preference data using on-policy trajectories guided by outcome and process signals. 
    \item \textbf{New State-of-the-art financial LLM} (\ourmodel) at the 8b parameter scale based on the above.
\end{itemize}
}



\section{Related Work}
\label{sec.related_work}

\begin{table*}[t]
\setlength{\tabcolsep}{2pt}
\centering
\resizebox{0.9\textwidth}{!}{
\begin{tabular}{l|l|l|l|l}
\toprule
\multirow{2}{*}{\textbf{\begin{tabular}[c]{@{}c@{}}Finance \\ LLM\end{tabular}}} & \multirow{2}{*}{\textbf{Capabilities}} & \multicolumn{2}{c|}{\textbf{Recipe}} &  \multicolumn{1}{c}{\multirow{2}{*}{\textbf{Evaluation}}} \\
 &  & \multicolumn{1}{c|}{\textbf{Model Recipe}} & \multicolumn{1}{c|}{\textbf{Data Recipe}} &  \\
 \toprule
AdaptLLM & Concept & CPT & \textbf{CPT}: Financial text + heuristic QAs constructed from the text &   Financial tasks + Direct answer  \\
\hline
PIXIU & Task & IT & \textbf{IT}: Financial tasks & Financial tasks + Direct answer \\
\hline
FinLLM & Concept, Task & CPT → IT & \textbf{CPT}: Financial text + Fineweb; \textbf{IT}: Filtered Financial tasks &  Financial tasks + Direct answer   \\
\hline
FinTral & Concept, Task & CPT → IT → PA & \textbf{CPT}: Financial text;  \textbf{IT}: Financial tasks; \textbf{PA}: Outcome signal only&  Financial tasks + Direct answer  \\
\hline
Palmyra-Fin & \multicolumn{4}{c}{SoTA public checkpoint, but recipe is not disclosed} \\
\hline
\rowcolor[HTML]{D7F6FF} 
\ourmodel & \begin{tabular}[c]{@{}c@{}}Concept, IF/Chat,\\ Task, Reasoning\end{tabular} & CPT+IT → PA & \begin{tabular}[l]{@{}l@{}}\textbf{CPT}: Financial + General text. \\ \textbf{IT}: Financial + General tasks\\ \textbf{PA}: A novel PA that leverages outcome and process signals\end{tabular} & \begin{tabular}[l]{@{}l@{}}General + Financial tasks; Similar + Novel tasks\\ Knowledge Recall + Reasoning tasks \\ Direct answer + CoT \end{tabular} \\
\bottomrule
\end{tabular}
}
\vspace{-3mm}
\caption{\small Comparison between \ourmodel with other finance LLMs.} 
\vspace{-3mm}
\label{tab.fin_compare}
\end{table*}

\noindent\textbf{Finance LLMs}
{Table~\ref{tab.fin_compare} summarizes popular finance-specific LLMs developed through domain-adaptive post-training.  AdaptLLM~\citep{cheng2024adapting} focuses on CPT and constructs heuristic QA tasks from raw text, but it considers only five financial end tasks. PIXIU~\citep{xie2023pixiulargelanguagemodel} focus on instruction-following by creating a financial instruction-tuning dataset from diverse open financial tasks and designing a benchmark with nine end tasks for evaluation. FinLLM~\citep{xie2024openfinllmsopenmultimodallarge} extends post-training across multiple stages, first performing CPT, then IT, and incorporating multi-modal capabilities via IT. It includes some general-domain data (e.g., FineWeb~\citep{penedo2024finewebdatasetsdecantingweb}) but does not explore its impact systematically. Following this line, FinTral~\citep{bhatia2024fintralfamilygpt4level} is the only open FinLLM to include PA, where preference labels were given by GPT-4 on the final outcome, considering only coarse-grained signals. It also introduces multi-modality via IT and integrates tool use and retrieval in PA training. Additionally, Palmyra-Fin~\citep{Palmyra-Fin-70B-32k}, a recent state-of-the-art FinLLM, reports high performance on finance tasks, particularly CFA exams\footnote{\url{https://www.cfainstitute.org/programs/cfa-program/exam}}, but its training recipe remains undisclosed. 

Comparing to \ourframework, none of these models explicitly identify \textit{desirable capabilities} as we do with \textit{FinCap}, nor do they systematically explore trade-offs between CPT, IT and PA to develop a more effective training recipe. They also do not incorporate fine-grained process signals in PA to improve reasoning, as we do in \textit{FinRec}. Additionally, their evaluations lack the broader range of tasks, methods, and similarities, including reasoning tasks and CoT evaluations, that we adopt in \textit{FinEval}. Finally, unlike Palmyra-Fin, \ourmodel is fully open-source, ensuring complete transparency in its training recipe, datasets, and evaluation methods, while achieving SoTA in its size category.
} 

{\noindent\textbf{PA for reasoning} We explore training-time approaches for improving reasoning \citep{jiao-etal-2024-learning,deepseekr1}. These methods first collect trajectories 
and then train the LLM with the collected trajectories. This helps the model reason more accurately and faster during inference. }
To collect reasoning trajectories, there are two main approaches. The first is \textit{search-based} \citep{setlur2024rewardingprogressscalingautomated,snell2024scalingllmtesttimecompute}, where a trained Reward Model (RM) is used to guide a search method (e.g., Best-of-N, Beam Search) to identify the best reasoning path. The second is \textit{revision-based} \citep{bai2022constitutionalaiharmlessnessai,du2023improving,madaan2023selfrefine,saunders2022selfcritiquingmodelsassistinghuman}, which attempts to improve the generation distribution through multi-round interactions, often by leveraging feedback from itself or another strong LLM to refine the input prompt. In practice, revision-based methods have shown mixed results and have not yet been well established as reliable for achieving improvements \citep{huang2024large}. In contrast, search-based methods have been shown to be  more effective.
{In \ourframework, we propose a novel training-time method that leverages a search-based trajectory collection approach, incorporating both outcome and process rewards from a Generative RM (GenRM).}


\section{\ourframework Framework}


In \ourframework, we first identify \textit{four} desired capabilities for a finance-expert LLM (\textbf{FinCap}, \S\ref{sec.perspectives}). We then develop the training recipe \textbf{FinRec},  which includes both the \textit{model recipe} that performs CPT and IT jointly followed by PA, and the \textit{data recipe}, which examines the impact of in-domain, general-domain, and mixed-domain data while introducing a novel data construction approach for PA (\S\ref{sec.model_train}). We then introduce \textbf{FinTrain} (\S\ref{sec.fin_train}), a set of carefully curated training datasets designed to mitigate forgetting while effectively learning domain-specific knowledge. Finally, we propose an evaluation framework \textbf{FinEval} (\S\ref{sec.evalaution_framwork}), which considers a diverse set of tasks, ranging from familiar to novel and from general to domain-specific, while also evaluating both direct-answer and CoT methods.

\vspace{-0.2em}
\subsection{Core Capabilities (FinCap)}
\label{sec.perspectives}


We began by conducting a comprehensive survey of existing work and consulting two financial domain experts: a banking industry advisor and a financial industry product manager. From this, we identified four key fundamental capabilities essential for a finance LLM: understanding domain-specific concepts to process financial language accurately, performing domain-specific tasks to solve real-world problems, reasoning effectively to analyze complex financial data, and following instructions to interact naturally in practical applications. These capabilities are deeply interconnected: reasoning depends on conceptual knowledge, while instruction-following ensures effective communication.


\noindent\textbf{$\bullet$ Domain specific concepts.} A domain typically includes its own specific concepts. For example, `bond' in finance refers to a loan agreement between an investor and a borrower. Adapting the LLM to domain-specific concepts is crucial, as these concepts form the fundamental building blocks of domain knowledge. However, this adaptation should not come at the cost of losing knowledge about general concepts, which are essential for both domain-specific and general tasks.

\noindent\textbf{$\bullet$ Domain specific tasks.} While many NLP tasks, such as NER or sentiment analysis, are shared across different domains, a domain typically has its own tasks. For example, stock movement detection is primarily found in finance. Adapting LLMs to these domain-specific tasks is important, as it demonstrates how they can leverage domain-specific concepts to solve tailored tasks effectively.

\noindent\textbf{$\bullet$ Reasoning.} For complex tasks, reasoning with concepts is a highly desired capability in LLMs. For example, in finance, the LLM is often required to analyze a company's financial report, involving extensive reasoning, particularly mathematical reasoning, to compute key financial concepts such as market rate or earnings per share. 

\noindent\textbf{$\bullet$ Instruction-Following (IF) and chat. }This is a core capability for both general and domain-specific LLMs, as tasks are often presented in the form of instruction following or conversation. 


\vspace{-0.2em}
\subsection{\ourframework Training Recipe (FinRec)}
\label{sec.model_train}

As shown in Figure~\ref{fig.overview}, FinRec consists of two recipes: the \textit{model recipe}, which focuses on the training stages and losses, and the \textit{data recipe}, which focuses on constructing training data.

\subsubsection{Model Recipe}

Previous studies often de facto treat domain-adaptive post-training as a sequential process involving, or partially involving, CPT, IT, and PA. However, our experiments with LLaMA3-8B-Inst show key trade-offs among these stages (App. \ref{sec.ablation}). While CPT is effective at introducing domain concepts, it often leads to \textit{significant forgetting} of general concepts and instruction-following capabilities.  In contrast, IT strengthens instruction-following capabilities and introduces domain-specific tasks with minimal forgetting. IT alone however struggles with \textit{task generalization}. PA is effective for learning reasoning but depends heavily on high-quality preference data, which can be difficult to synthesize. To address these limitations, we propose a \textit{joint CPT+IT approach}, resulting in {CPT+IT} checkpoint. Subsequently, PA is performed with a novel trajectory collection method that provides fine-grained supervision signals.

\noindent\textbf{Joint continual pre-training and instruction-tuning (CPT + IT).}
In this stage, the goal is to learn domain-specific knowledge while maintaining general capabilities, such as instruction-following. It is well known that CPT can adapt the LLM to learn domain-specific concepts while IT can help learn the domain-specific and instruction-following tasks \cite{ke2022dga,wei2022finetuned}. Typically CPT involves next-token prediction \textit{without masking} any context tokens, and IT involves next-token prediction with \textit{instructions masked out}; thus training them sequentially from an instruction-tuned LLM naturally leads to forgetting general capabilities, including instruction-following. Intuitively, if the loss function incorporates both CPT and IT, forgetting can be largely mitigated \cite{scialom-etal-2022-fine}.\footnote{This is akin to ‘replay’ method in continual learning.} To achieve this, we mix CPT and IT data, effectively performing joint optimization, as the only difference between the two is whether the instruction is masked. This approach also facilitates knowledge transfer, as CPT helps the model learn domain knowledge, which can be leveraged by IT training. More importantly, since concepts {learned from CPT}
are often inherently more generalizable
due to the shared nature of concepts
across tasks, jointly training CPT and IT can improve generalization without require exposure to a diverse range of tasks 
, which is often impractical in
certain domains, particularly long-tail
ones. Since CPT datasets are typically much larger than IT datasets, we downsample CPT data to match the size of IT data, allowing for effective joint training.


\noindent\textbf{Improving reasoning with preference alignment.}
CPT+IT improves capabilities such as in general and domain-specific concepts, tasks and IF/Chat. However, we find that the resulting model lacks in its reasoning capability, especially when it comes to complex reasoning like solving problems in CFA exams, 
where it is important to make each reasoning step correct.
We use PA for this, which trains the model to assign higher probability mass to better generations, and has been shown to be effective in enhancing LLM reasoning capabilities \citep{lambert2024tulu3,jiao-etal-2024-learning}. Specifically, we employ Direct Preference Optimization or DPO \citep{rafailov2023direct}, which allows the model to learn from both positive and negative examples, providing a richer learning signal compared to SFT.  
We synthetically generate such data from the \textit{on-policy} model, i.e., the jointly trained CPT+IT checkpoint, as it has shown the strongest performance in preliminary experiments (Appx.~\ref{sec.joint_cpt_it}). We propose a novel trajectory collection method that provides fine-grained step-level supervision signals (\S\ref{sec.data_recipe})

\subsubsection{Data Recipe}
\label{sec.data_recipe}

While data quality and diversity are standard concerns in LLM training, we focus on two under-explored challenges: (a) the impact of in-domain, general-domain, and mixed-domain datasets on model performance at different training stages; (b) the generation of fine-grained supervision signals in PA to improve reasoning. 

\noindent\textbf{Mixture of in-domain and general-domain data.} 
{
Most existing finance LLMs rely exclusively on in-domain data in post-training with the exception of FinLLM, which uses general domain data in CPT (see Table~\ref{tab.fin_compare}). Intuitively, this exclusive reliance on in-domain data can lead to forgetting of general knowledge in the original pre-trained LLM. To understand how the forgetting happens across different stages, we conduct ablations by constructing three versions of data for \textbf{each training stage}: in-domain, general-domain, and a mixture of both. 
Experiments (App. \ref{sec.ablation}) show that the impact of forgetting \textbf{decreases progressively} from CPT to IT to PA, with CPT experiencing the most severe forgetting and PA the least. Guided by these findings, we adopt a mix of in- and general-domain data for CPT+IT training to maximize both specialization and retention of essential general knowledge.
}

\begin{figure}[t]
\centering
\includegraphics[width=\columnwidth]{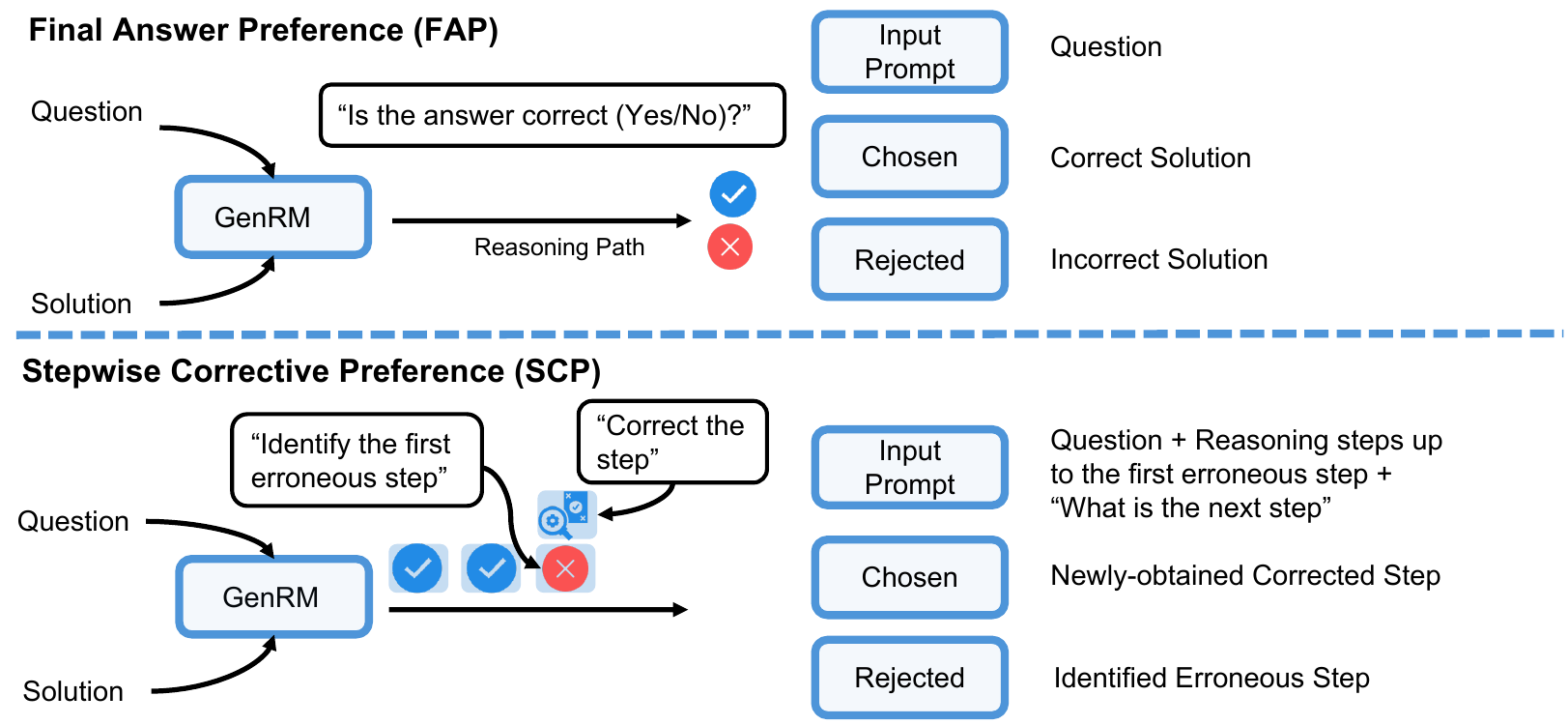}
\vspace{-8mm}
\caption{\small An overview of the proposed final answer preference (FAP) and stepwise corrective preference (SCP). In FAP, we collect trajectories from the GenRM by evaluating the entire solution.
In SCP, we collect trajectories from the GenRM, by identifying and correcting the first erroneous step. 
}
\label{fig.reasoning}
\vspace{-5mm}
\end{figure}

\noindent\textbf{Preference data construction for reasoning.}
Existing training methods to improve reasoning primarily rely on outcome-based rewards, which provide sparse supervision and do not guide intermediate reasoning steps. At the same time, stepwise reward models can be computationally expensive if applied at every step. To strike a balance, we employ a Generative Reward Model (GenRM) and design Final Answer Preference (FAP) to efficiently collect preference signals at the \textit{final answer level (outcome reward)}, while also collecting Stepwise Corrective Preference (SCP) at the reasoning step level (process reward) by asking the GenRM to identify and correct the first erroneous step. By combining these two complementary strategies, our PA provides stronger supervision signals for reasoning improvements while maintaining efficiency, making it particularly suitable for domains like finance, where both accuracy and efficiency are critical. Figure~\ref{fig.reasoning} illustrates the proposed method. 


\noindent\textbf{$\bullet$ Final Answer Preference (FAP).} Given a prompt and a model generated solution, we prompt the GenRM to give a holistic judgment for the entire solution using a single ``Yes'' or ``No'' token. 
We then use the correct solutions as chosen samples and the incorrect solutions as rejected samples.

\noindent\textbf{$\bullet$ Stepwise Corrective Preference (SCP).} Since reasoning could be complex (e.g., CFA exams) and \textit{process rewards} have been shown to be more effective in such cases~\citep{lightman2024lets}, we further leverage the GenRM to provide step-level signals. Instead of requesting rewards at each step, which has been shown  to be unnecessary in \cite{lightman2024lets,luo2024improvemathematicalreasoninglanguage}, we prompt the GenRM to identify the \textit{first} erroneous step and ask it to provide a correction for that step. 
Using this correction, we construct a preference data sample. The input prompt is formed by concatenating the original question, the candidate reasoning steps up to the first error, and a follow-up question framed as ``What is the next step?''. The \textit{chosen} response of this preference sample is the newly-obtained corrected step, while the original first erroneous step is deemed as \textit{rejected} response. This approach produces trajectories that focus on predicting the correct next step given a reasoning prefix, unlike FAP, which requires a prediction of the entire reasoning trajectory (see App.~\ref{sec.gen_rm_detail} for prompt details).


\subsection{\ourframework Training Data (FinTrain)}
\label{sec.fin_train}




In FinTrain, we carefully balance the trade-off between quality and quantity of the training data  at each stage. Specifically, in CPT, we leverage available general-domain supervised data, like NaturalInstrution \cite{naturalinstructions}.
Since such data has been carefully curated and cleaned for labeling, they maintain good quality. 
For {quantity and diversity}, which is essential for learning new domain knowledge during CPT, we collect large-scale, diverse data from relevant sources, including 70 financial websites and books covering 12 financial topics, like CFA exam preparation materials. We further use a strong LLM to filter out low-quality tasks based on an additive scale prompt \cite{yuan2024selfrewarding}. {This results in approximately 6B tokens.}

For IT, to promote diversity, we conduct a broad survey and source general, financial, instruction-following, and reasoning datasets from public datasets. We also include large open QA datasets like FinQA \cite{chen-etal-2021-finqa}. To ensure quality, we prioritize datasets that shown to perform well in the literature, like UltraChat \cite{ding2023enhancing}. We also incorporate exercises or demonstrations from books that often contain human-written CoT. {The final IT dataset consists of $\sim$3M prompts.}

For PA, we use CFA preparation materials as a representative source for in-domain reasoning as they cover diverse financial scenarios, emphasize complex reasoning, and, most importantly, are derived from real-world exams. We construct preference data with FAP and SCP introduced in \S\ref{sec.data_recipe}. {The final PA dataset consists of about 32K prompts.} Additional details are given in App. \ref{sec.ablation_fin_train}.


\subsection{\ourframework Evaluation (FinEval)}
\label{sec.evalaution_framwork}

Our evaluation framework \textbf{FinEval} is designed to systematically assess model performance across  unseen tasks. Unlike prior studies that rely on a narrow set of domain-specific tasks, FinEval categorizes tasks by similarity (similar vs. novel tasks), domain specificity (general vs. domain-specific vs. reasoning tasks), and evaluation methods (direct-answer vs. chain-of-thought). By structuring evaluations along these dimensions, FinEval {consists of \textbf{35} tasks and} can serve as a comprehensive benchmark for the expected capabilities going beyond simple task-based evaluation. {\color{black}We took extra care to ensure that FinEval does not duplicate any samples from FinTrain: the 10-gram contamination rate is only \textbf{0.003\%}, indicating minimal overlap (see  App.~\ref{ap.contamination})}. We provide details about the evaluation tasks and methods in App. \ref{sec.ablation_fin_eval}. 

\section{Experiments}
\label{sec.experiment}


\begin{table}[t]
\centering
\setlength{\tabcolsep}{2pt}
\resizebox{\columnwidth}{!}{
\begin{tabular}{ll>{\columncolor{lightblue}}lll|lll>{\columncolor[gray]{0.8}}l}
\toprule
 Task & Benchmark & \begin{tabular}[c]{@{}l@{}}\ourmodel \\ 8B\end{tabular} & \begin{tabular}[c]{@{}l@{}}Llama3 \\ Instruct \\ 8B \end{tabular} & \begin{tabular}[c]{@{}l@{}}Llama3.1 \\Instruct\\ 8B \end{tabular} & \begin{tabular}[c]{@{}l@{}}Palmyra\\Fin \\ 70B\end{tabular} & \begin{tabular}[c]{@{}l@{}}Phi\\3.5-mini \\ Instruct\\ 3.8B\end{tabular} & \begin{tabular}[c]{@{}l@{}}Mistral\\Nemo \\ instruct \\ 12B\end{tabular} & GPT4o \\
\toprule 
 Sentiment Ana. & FPB \small{(Acc)} & \underline{\textbf{91.13}}$^\checkmark$ & 73.09 & 71.55 & 67.11 & 78.04 & 78.25 & 82.16 \\
 Sentiment Ana. & FiQA SA \small{(Acc)} & \underline{\textbf{95.32}}$^\checkmark$ & 77.87 & 70.64 & 71.91 & 69.36 & 55.74 & 68.51 \\
 Monetary Policy & FOMC \small{(Acc)} & \textbf{64.31}$^\checkmark$ & 56.65 & 54.64 & 63.10 & 58.47 & 57.86 & \underline{67.94} \\
Named Entity & NER \small{(Rouge1)} & \underline{\textbf{76.69}}$^\checkmark$ & 45.03 & 51.22 & 54.29 & 39.37 & 49.84 & 43.02 \\
Abs Summ. & EDTSUM \small{(Rouge1)} & \underline{\textbf{53.78}}$^\checkmark$ & 11.50 & 12.53 & 21.77 & 19.97 & 12.32 & 18.15 \\
\bottomrule
\end{tabular}
}
 \vspace{-3mm}
\caption{\small Results on \textbf{similar (unseen)} tasks. \ourmodel is \colorbox{lightblue}{highlighted in blue} while the closed model is \colorbox[gray]{0.8}{highlighted in gray}. The best performing model for 8b on each benchmark is \textbf{bolded}. The overall best performance across all models is \underline{underlined}. $^\checkmark$ indicates that \ourmodel outperforms the base Llama3-8b-inst. 
}
\vspace{-6mm}
\label{tab.similar_results}
\end{table}

\begin{table*}[t]
\centering
\setlength{\tabcolsep}{2pt}
\resizebox{0.8\textwidth}{!}{
\begin{tabular}{llll>{\columncolor{lightblue}}lll|lll>{\columncolor[gray]{0.8}}l}
\toprule
Capability & Domain & Task & Benchmark & \begin{tabular}[c]{@{}l@{}}\ourmodel \\ 8B\end{tabular} & \begin{tabular}[c]{@{}l@{}}Llama3 \\ Instruct \\ 8B \end{tabular} & \begin{tabular}[c]{@{}l@{}}Llama3.1 \\Instruct\\ 8B \end{tabular} & \begin{tabular}[c]{@{}l@{}}Palmyra\\Fin \\ 70B\end{tabular} & \begin{tabular}[c]{@{}l@{}}Phi\\3.5-mini \\ Instruct\\ 3.8B\end{tabular} & \begin{tabular}[c]{@{}l@{}}Mistral\\Nemo \\ instruct \\ 12B\end{tabular} & GPT4o \\
\toprule 
\textbf{Concept} & General & Knowledge Recall & MMLU \small{(CoT, Acc)} & 47.42 & \textbf{48.14} & 47.42 & 54.93 & 45.07 & 49.64 & \underline{63.88} \\
& & & AI2-ARC \small{(CoT, Acc)} & 89.43$^\checkmark$  & 89.29 & \textbf{89.80} & 89.01 & 87.25 & 88.19 & \underline{97.85} \\
 &  &  & Nq-open \small{(CoT, Acc)} & 19.20$^\checkmark$  & 18.47 & \textbf{22.52} & 19.25 & 6.20 & 17.01 & \underline{27.92} \\
  & Finance & Knowledge Recall & MMLU-Finance \small{(Acc)} & 64.20 & 65.71 & \textbf{66.74} & 75.15 & 68.17 & 61.88 & \underline{86.52} \\
   \hline
\textbf{Task} & Finance & Extractive Summ. & Flare-ECTSUM \small{(Rouge1)} & 34.10 & \textbf{35.92} & 35.77 & 33.24 & 35.52 & \underline{37.86} & 35.90 \\
 &  & ESG Issue & MLESG \small{(Acc)} & \textbf{40.67}$^\checkmark$ & 36.33 & 36.00 & 39.67 & 38.33 & 32.67 & \underline{45.67} \\
 &  & Rumor Detection & MA \small{(Acc)} & 84.00$^\checkmark$ & 82.60 & \textbf{84.20}  & 62.60 & 75.40 & \underline{85.20} & 73.80 \\
 &  & Stock Movement & SM-Bigdata \small{(CoT, Acc)} & 54.14 & \textbf{55.3} & 46.06 & 48.70 & 53.26 & 53.53 & 49.18 \\
 &  &  & SM-ACL \small{(CoT, Acc)} & \textbf{51.99}$^\checkmark$ & 50.51 & 45.30 &  51.21 & 49.84 & 50.75 & 50.97 \\
 &  & & SM-CIKM \small{(CoT, Acc)} & 54.94 & \textbf{55.56} & 48.03 & 52.92 & 50.03 & 53.28 & 49.78 \\
 &  & Fraud Detection & CRA-CCF \small{(CoT, Mcc)} & 0.83$^\checkmark$ & -0.32 & \textbf{2.73} & 3.12 & 1.20 & 3.94 & 6.16 \\
 &  & & CRA-CCFraud \small{(CoT, Acc)} & \textbf{34.03}$^\checkmark$ & 14.78 & 17.3 & 33.03 & 45.33 & 32.94 & \underline{49.57}  \\
 &  & Credit Scoring & Flare-German \small{(CoT, Acc)} & \textbf{64.00}$^\checkmark$ & 33.50 & 15.00 & 12.00 & 49.50 & 32.50 & 17.00 \\
 &  & & Flare-Astralian \small{(CoT, Acc)} & 44.60 & \textbf{66.91} & 11.51 & 12.95  & 46.76 & 56.12 & 51.80 \\
 &  & & CRA-LendingClub \small{(CoT, Acc)} & \textbf{68.49}$^\checkmark$ & 52.69 & 25.38 &  23.40 & 48.87 & 21.03 & 65.03 \\
 &  & Distress Ident.  & CRA-Polish \small{(CoT, Mcc)} & \textbf{15.30}$^\checkmark$ & 12.37 & 15.07 &  13.78  & 69.14 & 11.18 & 17.38 \\
 &   & & CRA-Taiwan \small{(CoT, Acc)} & \textbf{40.81}$^\checkmark$ & 12.01 & 35.97 & 52.58  & 69.96 & 57.88 & 8.57 \\
 &  & Claim Analysis & CRA-ProroSeguro \small{(CoT, Acc)} & 35.14 & \textbf{96.98} & 44.33 & 56.20  & 25.86 & 32.58 & 96.60 \\
 &  &  & CRA-TravelInsurance \small{(CoT,Acc)} & 41.52$^\checkmark$ & 6.39 & 80.31  & 17.28 & \underline{\textbf{94.48}} & 73.64 & 54.03 \\
  &  & Tabular QA & *Flare-TATQA \small{(CoT, Acc)} & \textbf{66.61}$^\checkmark$ & 63.43 & 63.70 & 64.21 & 57.70 & 66.40 & \underline{74.90}  \\
 &  & Open QA & *Finance Bench \small{(CoT, Acc)} & \textbf{54.00}$^\checkmark$ & 52.70 & 38.00 & \underline{{56.67}} & 40.70 & 55.30 & 51.30  \\
 \hline
 \textbf{IF/Chat} & General & Precise IF & MT-bench \small{(1,2 turn avg)} & 7.36 & 7.88 & \textbf{7.92} & 5.80 & 8.38 & 7.84 & \underline{9.10} \\
\hline
\textbf{Reasoning} & Math & Math Reasoning & MathQA \small{(CoT, Acc)} & \textbf{55.08}$^\checkmark$ & {51.16} & {49.35} & {41.51} & {39.40} & {52.46} & \underline{70.82} \\
 & General & Social Reasoning & Social-IQA \small{(CoT, Acc)} & {\textbf{75.23}}$^\checkmark$ & {68.83} & {70.73} & 77.28 & {72.82} & {62.95} & \underline{78.92} \\
 &  & Common Sense & Open-book-qa \small{(CoT, Acc)} & {\textbf{82.60}}$^\checkmark$ & {77.00} & {82.20} & 87.00 & {80.20} & {76.40} & \underline{94.60} \\
 &  &  & Hellaswag \small{(CoT, Acc)} & \underline{\textbf{81.90}}$^\checkmark$ & {73.34} & {69.10} & {69.69} & {67.89} & {61.74} & {81.76}  \\
 &  &  & Winogrande \small{(CoT, Acc)} & {\textbf{70.32}}$^\checkmark$ & {62.51} & {66.69} & {74.27} & {72.22} & {65.82} & \underline{85.71} \\
 &  &  & PIQA \small{(CoT, Acc)} & {\textbf{85.85}}$^\checkmark$ & {79.82} & {81.45} & {86.72} & {82.05} & {77.91} & \underline{94.34} \\
 & Finance & Exam & CFA-Easy \small{(CoT, Acc)} & {\textbf{66.28}}$^\checkmark$ & {60.56} & {60.47} & {36.05} & {61.24} & {65.89} & \underline{83.14} \\
 &  &  & CFA-Challnge \small{(CoT, Acc)} & {\textbf{55.56}}$^\checkmark$ & {34.44} & {35.56} & {25.56} & {48.89} & {43.33} & \underline{74.44} \\
\bottomrule
\end{tabular}
}
 \vspace{-3mm}
\caption{\small Results on the \textbf{novel} tasks. The notations are the same as in  Table~\ref{tab.similar_results}. `*' indicates that `GPT4o' is used as the judge. 
`Mcc' refers to Matthews correlation coefficient, usually used in highly imbalanced data \citep{xie2024openfinllmsopenmultimodallarge}. 
}
\vspace{-6mm}
\label{tab.novel_results}
\end{table*}

We apply our method to the instruction-tuned Llama3-8b-inst, resulting in \textbf{\ourmodel} {\color{black}(GPT-4o is used as GenRM).} A summary of the hyper-parameters and computational resource requirements is given in Table~\ref{tab.final_recipe}. For evaluation, we compare  \ourmodel with a wide range of baselines models, including its base model, Llama3-8b-inst, and the 8B peer, Llama3.1-8b-inst. We also include comparisons with models of other sizes, such as Phi-3.5-mini-instruct~\citep{abdin2024phi3technicalreporthighly} (3.8B), and Mistral-Nemo-inst~\citep{jiang2023mistral} (12B), as well as the \textit{closed model} GPT-4o~\citep{aaa+23}. Furthermore, we evaluated against the latest SoTA finance-specific LLM, Palmyra-Fin (70B)~\citep{Palmyra-Fin-70B-32k}. Note that there are other financial LLMs available, such as FinMa~\citep{xie2023pixiulargelanguagemodel}
and FinLLaVA~\citep{xie2024openfinllmsopenmultimodallarge}. 
However, they are either not publicly available (FinLLaVA) or based on less advanced LLMs (e.g., LLaMA2). In preliminary experiments, these models performed considerably worse than our model {\color{black}(see App.~\ref{ap.prelim_baseline})}. Therefore, we have only included the SoTA financial LLM in our comparisons. \\



\subsection{Main Results}


\noindent\textbf{Similar (unseen) tasks.} To validate our approach, we first evaluate \ourmodel on tasks that are similar (yet unseen) to the tasks used for training 
{(e.g., test task EDTSUM (abstractive summarization) is similar to the training task TradeTheEvent (abstractive summarization))}.
From Table \ref{tab.similar_results}, we observe that \ourmodel outperforms all other baselines in its size category by 10\% - 25\% absolute gain. It also surpasses significantly larger models, such as the finance-specific Palmyra-Fin (70B). Notably, \ourmodel also exceeds the performance of GPT-4o. These results are not very surprising since the test tasks are not entirely novel, but it demonstrates the effectiveness of our data and model recipe for domain-adaptive post-training.

\noindent\textbf{Novel tasks.} We now evaluate the generalization of \ourmodel on the completely novel tasks that are also aligned to the expected capabilities (FinCap). Table \ref{tab.novel_results} presents the results.  
Below, we summarize the key takeaways from the comparison:

\noindent\textbf{$\bullet$ \ourmodel preserves general concepts (rows 2-5).} We observe that \ourmodel performs better or remains competitive with its base model in general knowledge recall tasks, indicating that it effectively preserves general capabilities and mitigating forgetting. It performs slightly worse than the base model in finance knowledge recall (MMLU-Finance), despite our finding that the CPT benefits IT (see ablations in Appendix \ref{sec.joint_cpt_it}). We hypothesize that CPT helps learn concepts that are helpful but differ from those emphasized in MMLU-Finance.

\noindent\textbf{$\bullet$ \ourmodel is effective in the majority of tasks (rows 6-22).} It outperforms the base model in 13 out of 17 tasks, demonstrating that our approach can lead to models that generalize well to novel, unseen tasks requiring the same capabilities. 

\noindent\textbf{$\bullet$ \ourmodel preserves IF/Chat capabilities (row 23).} 
\ourmodel achieves a competitive MT-Bench score compared to the base model, indicating that it effectively maintains the IF capability.

\noindent\textbf{$\bullet$ \ourmodel excels in reasoning tasks (rows 24-31).} For reasoning capability, \ourmodel significantly outperforms the base models across all considered benchmarks in a large amount (up to 20\% in CFA-Challenge), indicating substantial improvements in reasoning capability.



\subsection{Further Analysis and Ablations}

As discussed in \S\ref{sec.model_train}, we performed a number of data and model ablations in pursuit of designing the best training recipe (including parameter-efficient finetuning methods like LoRA) for \ourmodel. Those ablations are detailed in Appendix \ref{sec.ablation}. In this section, we present the impact of our PA strategy in the overall post-training process.

\begin{table}[t]
\centering
\setlength{\tabcolsep}{2pt}
\resizebox{0.75\columnwidth}{!}{
\begin{tabular}{ll>{\columncolor{lightblue}}l>{\columncolor{lightblue}}l}
\toprule
 Task & Benchmark & \ourmodel & \begin{tabular}[c]{@{}l@{}} \ourmodel \\ (w/o PA)\end{tabular}  \\
\toprule
Sentiment Ana. & FPB  & 91.13 & {\textbf{92.99}} \\
Sentiment Ana. & FiQA SA & \textbf{95.32} & 94.47 \\
Monetary Policy & FOMC  & {\textbf{64.31}} & 63.10 \\
Named Entity & NER  & {\textbf{76.69}} & 74.33  \\
Abs. Summ. & EDTSUM & 53.78 & {\textbf{54.21}} \\
\bottomrule
\end{tabular}
}
 \vspace{-3mm}
\caption{\small Ablation on PA on \textbf{similar (unseen)} evaluation set. 
}
\vspace{-6mm}
\label{tab.similar_result_abalation}
\end{table}

\begin{table}[t]
\centering
\setlength{\tabcolsep}{2pt}
\resizebox{\columnwidth}{!}{
\begin{tabular}{llll>{\columncolor{lightblue}}l>{\columncolor{lightblue}}l}
\toprule
Capability & Domain & Task & Benchmark & \ourmodel 8B & \begin{tabular}[c]{@{}l@{}} \ourmodel \\ (w/o PA) \end{tabular}  \\
\toprule
\textbf{Concept} & General & Knowledge Recall & MMLU  & \textbf{47.42} & 47.22 \\
& & & AI2-ARC  & \textbf{89.43} & 88.95  \\
 &  &  & Nq-open  & \textbf{19.20} & 16.20 \\
  & Finance & Knowledge Recall & MMLU-Finance  & \textbf{64.20} & 63.93 \\
   \hline
\textbf{Task} & Finance & Extract Summ. & Flare-ECTSUM & 34.10 & \textbf{34.41} \\
 &  & ESG Issue & MLESG  & 40.67 & \textbf{42.00}\\
 &  & Rumor Detection & MA  & 84.00 & \textbf{84.60}  \\
 &  & Stock Movement & SM-Bigdata  & \textbf{54.14} & 52.04  \\
 &  &  & SM-ACL  & \textbf{51.99} & 49.89  \\
 &  & & SM-CIKM  & \textbf{54.94} & 44.88  \\
 &  & Fraud Detection & CRA-CCF  & \textbf{0.83} & 0.61  \\
 &  & & CRA-CCFraud  & \textbf{34.03} & 32.32  \\
 &  & Credit Scoring & Flare-German  & \textbf{64.00} & 60.50 \\
 &  & & Flare-Astralian  & 44.60 & \textbf{51.80} \\
 &  & & CRA-LendingClub  & \textbf{68.49} & 65.96 \\
 &  & Distress Ident. & CRA-Polish  & \textbf{15.30} & 0.65 \\
 &   & & CRA-Taiwan  & 40.81 & \textbf{96.41} \\
 &  & Claim Analysis & CRA-ProroSeguro  & 35.14 & \textbf{86.57} \\
 &  &  & CRA-TravelInsurance & 41.52 & {\textbf{98.50}}  \\
  &  & Tabular QA & *Flare-TATQA  & \textbf{66.61} & 66.43  \\
 &  & Open QA & *Finance Bench  & \textbf{54.00} & 52.00  \\
 \hline
 \textbf{IF/Chat} & General & Precise IF & MT-bench & \textbf{7.36} & 7.29 \\
\hline
\textbf{Reasoning} & Math & Math Reasoning & MathQA  & \textbf{55.08} & {54.30}  \\
 & General & Social Reasoning & Social-IQA  & {\textbf{75.23}} & {73.64} \\
 &  & Common Sense & Open-book-qa  & {\textbf{82.60}} & {79.20} \\
 &  &  & Hellaswag  & {\textbf{81.90}} & {78.92}  \\
 &  &  & Winogrande  & {\textbf{70.32}} & {67.48}  \\
 &  &  & PIQA  & {\textbf{85.85}} & {84.39}  \\
 & Finance & Exam & CFA-Easy  & {\textbf{66.28}} & {62.31}  \\
 &  &  & CFA-Challnge  & {\textbf{55.56}} & {35.56} \\
\bottomrule
\end{tabular}
}
 \vspace{-3mm}
\caption{\small Abaltion on PA on \textbf{novel} evaluation set.
}
\vspace{-6mm}
\label{tab.novel_result_abalation}
\end{table}

Table \ref{tab.similar_result_abalation} presents the effectiveness of PA on similar tasks.  
We see that PA leads to improved performance in 3 out of 5 tasks, while not causing any significant forgetting on the other two. This is expected as PA primarily targets the reasoning tasks whereas these tasks do not need much reasoning. 

In Table \ref{tab.novel_result_abalation}, we show the same ablation for the novel tasks. 
In \textbf{Concept (rows 2-5)} and \textbf{IF/chat (row 23)} capabilities, removing PA often leads to worse results, indicating its effectiveness. In \textbf{Task (rows 6-22)}, we see a mixed performance with and without PA. This is again not surprising as PA specifically focuses on reasoning tasks. Interestingly, we observe that for certain tasks (e.g., CRA-TravelInsurance, CRA-Taiwan and CRA-ProroSeguro), PA negatively impacts performance, resulting in worse outcomes compared to without PA. Even GPT-4o performs poorly in these tasks.  This suggests that for some tasks, leveraging reasoning capabilities might not be beneficial, as these tasks could be inherently ``easy'' and solvable without the need for explicit reasoning. \textcolor{black}{Such observations align with prior findings  \citep{sprague2024cotcotchainofthoughthelps,liu2024mindstepbystep}.} In \textbf{Reasoning (rows 24-31)}, \ourmodel is significantly better than without PA variant, further confirming that our proposed FAP and SCP are particularly effective in improving reasoning performance beyond the already strong checkpoint of \ourmodel (w/o PA).
\section{Conclusion}

{\color{black}We introduce \ourframework, an open SoTA finance-specific post-training framework, consists of \textit{FinCap} that identifies four key capabilities; \textit{FinRec} which jointly trains CPT and IT, and constructing PA preference data with stepwise signals; \textit{FinTrain} that implements FinRec; and \textit{FinEval}, a comprehensive evaluation setup.
Under \ourframework, we develop \ourmodel, a SoTA finance LLM. In this development, we conduct a systematic study on effectively adapting a target domain through post-training. For each stage, we reveal the distinct challenges, objectives, and effective strategies. Looking ahead, we aim to scale up the base LLM and explore additional domain-specific capabilities using \ourframework.}


\section{Limitations}

While the recipe for \ourframework and \ourmodel are effective, the performance on novel unseen tasks still requires further improvement. For example, selectively employing reasoning capabilities only for questions that require such advanced reasoning might give better results.
Additionally, the data recipe is currently based on full-scale empirical experiments, which can be time-intensive. Developing low-cost experiments to reliably indicate the effectiveness of data in post-training could streamline this process and accelerate the development iteration. \textcolor{black}{It is also worth noting that the same recipe may not generalize well to other model families. Different architectures or pretraining strategies might require tailored recipe to achieve optimal results, emphasizing the need for adaptability in recipe design in future research.}
Finally, while we focus on the four key capabilities in finance, we acknowledge there could be additional requirements (e.g., multi-modality and sensitivity, see details in Appendix  \S\ref{sec.extra_capability}), and leave them for future work. 

\bibliography{custom}

\begin{thebibliography}{101}
\providecommand{\natexlab}[1]{#1}

\bibitem[{Abdin et~al.(2024)}]{abdin2024phi3technicalreporthighly}
Marah Abdin et~al. 2024.
\newblock \href {https://arxiv.org/abs/2404.14219} {Phi-3 technical report: A highly capable language model locally on your phone}.
\newblock \emph{Preprint}, arXiv:2404.14219.

\bibitem[{Alvarado et~al.(2015)Alvarado, Cesar, Verspoor, and Baldwin}]{salinas-alvarado-etal-2015-domain}
Salinas Alvarado, Julio Cesar, Karin Verspoor, and Timothy Baldwin. 2015.
\newblock \href {https://aclanthology.org/U15-1010} {Domain adaption of named entity recognition to support credit risk assessment}.
\newblock In \emph{Proceedings of the Australasian Language Technology Association Workshop 2015}, pages 84--90, Parramatta, Australia.

\bibitem[{Amini et~al.(2019{\natexlab{a}})Amini, Gabriel, Lin, Koncel-Kedziorski, Choi, and Hajishirzi}]{amini2019mathqa}
Aida Amini, Saadia Gabriel, Peter Lin, Rik Koncel-Kedziorski, Yejin Choi, and Hannaneh Hajishirzi. 2019{\natexlab{a}}.
\newblock \href {https://arxiv.org/abs/1905.13319} {Mathqa: Towards interpretable math word problem solving with operation-based formalisms}.
\newblock \emph{Preprint}, arXiv:1905.13319.

\bibitem[{Amini et~al.(2019{\natexlab{b}})Amini, Gabriel, Lin, Koncel-Kedziorski, Choi, and Hajishirzi}]{amini-etal-2019-mathqa}
Aida Amini, Saadia Gabriel, Shanchuan Lin, Rik Koncel-Kedziorski, Yejin Choi, and Hannaneh Hajishirzi. 2019{\natexlab{b}}.
\newblock \href {https://doi.org/10.18653/v1/N19-1245} {{M}ath{QA}: Towards interpretable math word problem solving with operation-based formalisms}.
\newblock In \emph{Proceedings of the 2019 Conference of the North {A}merican Chapter of the Association for Computational Linguistics: Human Language Technologies, Volume 1 (Long and Short Papers)}, pages 2357--2367, Minneapolis, Minnesota. Association for Computational Linguistics.

\bibitem[{Bach et~al.(2022)Bach, Sanh, Yong, Webson, Raffel, Nayak, Sharma, Kim, Bari, Fevry, Alyafeai, Dey, Santilli, Sun, Ben-David, Xu, Chhablani, Wang, Fries, Al-shaibani, Sharma, Thakker, Almubarak, Tang, Tang, Jiang, and Rush}]{bach2022promptsource}
Stephen~H. Bach, Victor Sanh, Zheng-Xin Yong, Albert Webson, Colin Raffel, Nihal~V. Nayak, Abheesht Sharma, Taewoon Kim, M~Saiful Bari, Thibault Fevry, Zaid Alyafeai, Manan Dey, Andrea Santilli, Zhiqing Sun, Srulik Ben-David, Canwen Xu, Gunjan Chhablani, Han Wang, Jason~Alan Fries, Maged~S. Al-shaibani, Shanya Sharma, Urmish Thakker, Khalid Almubarak, Xiangru Tang, Xiangru Tang, Mike Tian-Jian Jiang, and Alexander~M. Rush. 2022.
\newblock \href {https://arxiv.org/abs/2202.01279} {Promptsource: An integrated development environment and repository for natural language prompts}.
\newblock \emph{Preprint}, arXiv:2202.01279.

\bibitem[{Bai et~al.(2022)Bai, Kadavath, Kundu, Askell, Kernion, Jones, Chen, Goldie, Mirhoseini, McKinnon, Chen, Olsson, Olah, Hernandez, Drain, Ganguli, Li, Tran-Johnson, Perez, Kerr, Mueller, Ladish, Landau, Ndousse, Lukosuite, Lovitt, Sellitto, Elhage, Schiefer, Mercado, DasSarma, Lasenby, Larson, Ringer, Johnston, Kravec, Showk, Fort, Lanham, Telleen-Lawton, Conerly, Henighan, Hume, Bowman, Hatfield-Dodds, Mann, Amodei, Joseph, McCandlish, Brown, and Kaplan}]{bai2022constitutionalaiharmlessnessai}
Yuntao Bai, Saurav Kadavath, Sandipan Kundu, Amanda Askell, Jackson Kernion, Andy Jones, Anna Chen, Anna Goldie, Azalia Mirhoseini, Cameron McKinnon, Carol Chen, Catherine Olsson, Christopher Olah, Danny Hernandez, Dawn Drain, Deep Ganguli, Dustin Li, Eli Tran-Johnson, Ethan Perez, Jamie Kerr, Jared Mueller, Jeffrey Ladish, Joshua Landau, Kamal Ndousse, Kamile Lukosuite, Liane Lovitt, Michael Sellitto, Nelson Elhage, Nicholas Schiefer, Noemi Mercado, Nova DasSarma, Robert Lasenby, Robin Larson, Sam Ringer, Scott Johnston, Shauna Kravec, Sheer~El Showk, Stanislav Fort, Tamera Lanham, Timothy Telleen-Lawton, Tom Conerly, Tom Henighan, Tristan Hume, Samuel~R. Bowman, Zac Hatfield-Dodds, Ben Mann, Dario Amodei, Nicholas Joseph, Sam McCandlish, Tom Brown, and Jared Kaplan. 2022.
\newblock \href {https://arxiv.org/abs/2212.08073} {Constitutional ai: Harmlessness from ai feedback}.
\newblock \emph{Preprint}, arXiv:2212.08073.

\bibitem[{Bhatia et~al.(2024)Bhatia, Nagoudi, Cavusoglu, and Abdul-Mageed}]{bhatia2024fintralfamilygpt4level}
Gagan Bhatia, El~Moatez~Billah Nagoudi, Hasan Cavusoglu, and Muhammad Abdul-Mageed. 2024.
\newblock \href {https://arxiv.org/abs/2402.10986} {Fintral: A family of gpt-4 level multimodal financial large language models}.
\newblock \emph{Preprint}, arXiv:2402.10986.

\bibitem[{Bisk et~al.(2020)Bisk, Zellers, Bras, Gao, and Choi}]{Bisk2020}
Yonatan Bisk, Rowan Zellers, Ronan~Le Bras, Jianfeng Gao, and Yejin Choi. 2020.
\newblock Piqa: Reasoning about physical commonsense in natural language.
\newblock In \emph{Thirty-Fourth AAAI Conference on Artificial Intelligence}.

\bibitem[{Callanan et~al.(2024)Callanan, Mbakwe, Papadimitriou, Pei, Sibue, Zhu, Ma, Liu, and Shah}]{callanan-etal-2024-gpt}
Ethan Callanan, Amarachi Mbakwe, Antony Papadimitriou, Yulong Pei, Mathieu Sibue, Xiaodan Zhu, Zhiqiang Ma, Xiaomo Liu, and Sameena Shah. 2024.
\newblock \href {https://aclanthology.org/2024.finnlp-2.2} {Can {GPT} models be financial analysts? an evaluation of {C}hat{GPT} and {GPT}-4 on mock {CFA} exams}.
\newblock In \emph{Proceedings of the Eighth Financial Technology and Natural Language Processing and the 1st Agent AI for Scenario Planning}, Jeju, South Korea. -.

\bibitem[{Camburu et~al.(2018)Camburu, Rockt"{a}schel, Lukasiewicz, and Blunsom}]{NIPS2018_8163}
Oana-Maria Camburu, Tim Rockt"{a}schel, Thomas Lukasiewicz, and Phil Blunsom. 2018.
\newblock \href {http://papers.nips.cc/paper/8163-e-snli-natural-language-inference-with-natural-language-explanations.pdf} {e-snli: Natural language inference with natural language explanations}.
\newblock In S.~Bengio, H.~Wallach, H.~Larochelle, K.~Grauman, N.~Cesa-Bianchi, and R.~Garnett, editors, \emph{Advances in Neural Information Processing Systems 31}, pages 9539--9549. Curran Associates, Inc.

\bibitem[{Chen et~al.(2023{\natexlab{a}})Chen, Tseng, Kang, Lhuissier, Day, Tu, and Chen}]{chen-etal-2023-multi-lingual}
Chung-Chi Chen, Yu-Min Tseng, Juyeon Kang, Ana{\"\i}s Lhuissier, Min-Yuh Day, Teng-Tsai Tu, and Hsin-Hsi Chen. 2023{\natexlab{a}}.
\newblock Multi-lingual {ESG} issue identification.
\newblock In \emph{Proceedings of the Fifth Workshop on Financial Technology and Natural Language Processing and the Second Multimodal AI For Financial Forecasting}.

\bibitem[{Chen et~al.(2023{\natexlab{b}})Chen, Cano, Romanou, Bonnet, Matoba, Salvi, Pagliardini, Fan, K{\"o}pf, Mohtashami et~al.}]{chen2023meditron}
Zeming Chen, Alejandro~Hern{\'a}ndez Cano, Angelika Romanou, Antoine Bonnet, Kyle Matoba, Francesco Salvi, Matteo Pagliardini, Simin Fan, Andreas K{\"o}pf, Amirkeivan Mohtashami, et~al. 2023{\natexlab{b}}.
\newblock Meditron-70b: Scaling medical pretraining for large language models.
\newblock \emph{arXiv preprint arXiv:2311.16079}.

\bibitem[{Chen et~al.(2021)Chen, Chen, Smiley, Shah, Borova, Langdon, Moussa, Beane, Huang, Routledge, and Wang}]{chen-etal-2021-finqa}
Zhiyu Chen, Wenhu Chen, Charese Smiley, Sameena Shah, Iana Borova, Dylan Langdon, Reema Moussa, Matt Beane, Ting-Hao Huang, Bryan Routledge, and William~Yang Wang. 2021.
\newblock \href {https://doi.org/10.18653/v1/2021.emnlp-main.300} {{F}in{QA}: A dataset of numerical reasoning over financial data}.
\newblock In \emph{Proceedings of the 2021 Conference on Empirical Methods in Natural Language Processing}, pages 3697--3711, Online and Punta Cana, Dominican Republic. Association for Computational Linguistics.

\bibitem[{Chen et~al.(2022)Chen, Li, Smiley, Ma, Shah, and Wang}]{chen-etal-2022-convfinqa}
Zhiyu Chen, Shiyang Li, Charese Smiley, Zhiqiang Ma, Sameena Shah, and William~Yang Wang. 2022.
\newblock \href {https://doi.org/10.18653/v1/2022.emnlp-main.421} {{C}onv{F}in{QA}: Exploring the chain of numerical reasoning in conversational finance question answering}.
\newblock In \emph{Proceedings of the 2022 Conference on Empirical Methods in Natural Language Processing}, pages 6279--6292, Abu Dhabi, United Arab Emirates. Association for Computational Linguistics.

\bibitem[{Cheng et~al.(2024)Cheng, Huang, and Wei}]{cheng2024adapting}
Daixuan Cheng, Shaohan Huang, and Furu Wei. 2024.
\newblock \href {https://openreview.net/forum?id=y886UXPEZ0} {Adapting large language models via reading comprehension}.
\newblock In \emph{The Twelfth International Conference on Learning Representations}.

\bibitem[{Clark et~al.(2018)Clark, Cowhey, Etzioni, Khot, Sabharwal, Schoenick, and Tafjord}]{Clark2018ThinkYH}
Peter Clark, Isaac Cowhey, Oren Etzioni, Tushar Khot, Ashish Sabharwal, Carissa Schoenick, and Oyvind Tafjord. 2018.
\newblock Think you have solved question answering? try arc, the ai2 reasoning challenge.
\newblock \emph{ArXiv}, abs/1803.05457.

\bibitem[{Cobbe et~al.(2021)Cobbe, Kosaraju, Bavarian, Chen, Jun, Kaiser, Plappert, Tworek, Hilton, Nakano, Hesse, and Schulman}]{cobbe2021gsm8k}
Karl Cobbe, Vineet Kosaraju, Mohammad Bavarian, Mark Chen, Heewoo Jun, Lukasz Kaiser, Matthias Plappert, Jerry Tworek, Jacob Hilton, Reiichiro Nakano, Christopher Hesse, and John Schulman. 2021.
\newblock Training verifiers to solve math word problems.
\newblock \emph{arXiv preprint arXiv:2110.14168}.

\bibitem[{Colombo et~al.(2024{\natexlab{a}})Colombo, Pires, Boudiaf, de~Melo, Hautreux, Malaboeuf, Charpentier, Culver, and Desa}]{colombo2024saullmb}
Pierre Colombo, Telmo Pires, Malik Boudiaf, Rui Filipe Coimbra~Pereira de~Melo, Gabriel Hautreux, Etienne Malaboeuf, Johanne Charpentier, Dominic Culver, and Michael Desa. 2024{\natexlab{a}}.
\newblock \href {https://openreview.net/forum?id=NLUYZ4ZqNq} {Saul{LM}-54b \& saul{LM}-141b: Scaling up domain adaptation for the legal domain}.
\newblock In \emph{The Thirty-eighth Annual Conference on Neural Information Processing Systems}.

\bibitem[{Colombo et~al.(2024{\natexlab{b}})Colombo, Pires, Boudiaf, Culver, Melo, Corro, Martins, Esposito, Raposo, Morgado et~al.}]{colombo2024saullm}
Pierre Colombo, Telmo~Pessoa Pires, Malik Boudiaf, Dominic Culver, Rui Melo, Caio Corro, Andre~FT Martins, Fabrizio Esposito, Vera~L{\'u}cia Raposo, Sofia Morgado, et~al. 2024{\natexlab{b}}.
\newblock Saullm-7b: A pioneering large language model for law.
\newblock \emph{arXiv preprint arXiv:2403.03883}.

\bibitem[{DeepSeek-AI et~al.(2025)DeepSeek-AI, Yang, and et~al.}]{deepseekr1}
Daya~Guo DeepSeek-AI, Dejian Yang, and Li~et~al. 2025.
\newblock \href {https://arxiv.org/abs/2501.12948} {Deepseek-r1: Incentivizing reasoning capability in llms via reinforcement learning}.
\newblock \emph{Preprint}, arXiv:2501.12948.

\bibitem[{Ding et~al.(2023)Ding, Chen, Xu, Qin, Zheng, Hu, Liu, Sun, and Zhou}]{ding2023enhancing}
Ning Ding, Yulin Chen, Bokai Xu, Yujia Qin, Zhi Zheng, Shengding Hu, Zhiyuan Liu, Maosong Sun, and Bowen Zhou. 2023.
\newblock Enhancing chat language models by scaling high-quality instructional conversations.
\newblock \emph{arXiv preprint arXiv:2305.14233}.

\bibitem[{Du et~al.(2023)Du, Li, Torralba, Tenenbaum, and Mordatch}]{du2023improving}
Yilun Du, Shuang Li, Antonio Torralba, Joshua~B Tenenbaum, and Igor Mordatch. 2023.
\newblock Improving factuality and reasoning in language models through multiagent debate.
\newblock \emph{arXiv preprint arXiv:2305.14325}.

\bibitem[{Feng et~al.(2024)Feng, Dai, Huang, Zhang, Xie, Han, Chen, Lopez-Lira, and Wang}]{feng2024empoweringmanybiasingfew}
Duanyu Feng, Yongfu Dai, Jimin Huang, Yifang Zhang, Qianqian Xie, Weiguang Han, Zhengyu Chen, Alejandro Lopez-Lira, and Hao Wang. 2024.
\newblock \href {https://arxiv.org/abs/2310.00566} {Empowering many, biasing a few: Generalist credit scoring through large language models}.
\newblock \emph{Preprint}, arXiv:2310.00566.

\bibitem[{Gao et~al.(2024)Gao, Wettig, Yen, and Chen}]{gao2024trainlongcontextlanguagemodels}
Tianyu Gao, Alexander Wettig, Howard Yen, and Danqi Chen. 2024.
\newblock \href {https://arxiv.org/abs/2410.02660} {How to train long-context language models (effectively)}.
\newblock \emph{Preprint}, arXiv:2410.02660.

\bibitem[{Geva et~al.(2021)Geva, Khashabi, Segal, Khot, Roth, and Berant}]{geva2021strategyqa}
Mor Geva, Daniel Khashabi, Elad Segal, Tushar Khot, Dan Roth, and Jonathan Berant. 2021.
\newblock {Did Aristotle Use a Laptop? A Question Answering Benchmark with Implicit Reasoning Strategies}.
\newblock \emph{Transactions of the Association for Computational Linguistics (TACL)}.

\bibitem[{Gunasekar et~al.(2023)Gunasekar, Zhang, Aneja, Mendes, Giorno, Gopi, Javaheripi, Kauffmann, de~Rosa, Saarikivi, Salim, Shah, Behl, Wang, Bubeck, Eldan, Kalai, Lee, and Li}]{gunasekar2023textbooksneed}
Suriya Gunasekar, Yi~Zhang, Jyoti Aneja, Caio César~Teodoro Mendes, Allie~Del Giorno, Sivakanth Gopi, Mojan Javaheripi, Piero Kauffmann, Gustavo de~Rosa, Olli Saarikivi, Adil Salim, Shital Shah, Harkirat~Singh Behl, Xin Wang, Sébastien Bubeck, Ronen Eldan, Adam~Tauman Kalai, Yin~Tat Lee, and Yuanzhi Li. 2023.
\newblock \href {https://arxiv.org/abs/2306.11644} {Textbooks are all you need}.
\newblock \emph{Preprint}, arXiv:2306.11644.

\bibitem[{Gururangan et~al.(2020)Gururangan, Marasovi{\'c}, Swayamdipta, Lo, Beltagy, Downey, and Smith}]{gururangan-etal-2020-dont}
Suchin Gururangan, Ana Marasovi{\'c}, Swabha Swayamdipta, Kyle Lo, Iz~Beltagy, Doug Downey, and Noah~A. Smith. 2020.
\newblock \href {https://aclanthology.org/2020.acl-main.740} {Don{'}t stop pretraining: Adapt language models to domains and tasks}.
\newblock In \emph{Proceedings of the 58th Annual Meeting of the Association for Computational Linguistics}, Online. Association for Computational Linguistics.

\bibitem[{Hendrycks et~al.(2021)Hendrycks, Burns, Basart, Zou, Mazeika, Song, and Steinhardt}]{hendryckstest2021}
Dan Hendrycks, Collin Burns, Steven Basart, Andy Zou, Mantas Mazeika, Dawn Song, and Jacob Steinhardt. 2021.
\newblock Measuring massive multitask language understanding.
\newblock \emph{Proceedings of the International Conference on Learning Representations (ICLR)}.

\bibitem[{Hofmann(1994)}]{statlog_(german_credit_data)_144}
Hans Hofmann. 1994.
\newblock {Statlog (German Credit Data)}.
\newblock UCI Machine Learning Repository.
\newblock {DOI}: https://doi.org/10.24432/C5NC77.

\bibitem[{Hu et~al.(2021)Hu, Shen, Wallis, Allen-Zhu, Li, Wang, Wang, and Chen}]{hu2021loralowrankadaptationlarge}
Edward~J. Hu, Yelong Shen, Phillip Wallis, Zeyuan Allen-Zhu, Yuanzhi Li, Shean Wang, Lu~Wang, and Weizhu Chen. 2021.
\newblock \href {https://arxiv.org/abs/2106.09685} {Lora: Low-rank adaptation of large language models}.
\newblock \emph{Preprint}, arXiv:2106.09685.

\bibitem[{Huang et~al.(2024{\natexlab{a}})Huang, Chen, Mishra, Zheng, Yu, Song, and Zhou}]{huang2024large}
Jie Huang, Xinyun Chen, Swaroop Mishra, Huaixiu~Steven Zheng, Adams~Wei Yu, Xinying Song, and Denny Zhou. 2024{\natexlab{a}}.
\newblock \href {https://openreview.net/forum?id=IkmD3fKBPQ} {Large language models cannot self-correct reasoning yet}.
\newblock In \emph{The Twelfth International Conference on Learning Representations}.

\bibitem[{Huang et~al.(2024{\natexlab{b}})Huang, Cheng, Liu, Hao, Song, Xu, Yang, Liu, Zhang, Chai, Yuan, Zhang, Fu, Liu, Zhang, Wang, Qi, Xu, and Chu}]{Huang2024OpenCoderTO}
Siming Huang, Tianhao Cheng, Jason~Klein Liu, Jiaran Hao, Liuyihan Song, Yang Xu, J.~Yang, J.~H. Liu, Chenchen Zhang, Linzheng Chai, Ruifeng Yuan, Zhaoxiang Zhang, Jie Fu, Qian Liu, Ge~Zhang, Zili Wang, Yuan Qi, Yinghui Xu, and Wei Chu. 2024{\natexlab{b}}.
\newblock \href {https://arxiv.org/pdf/2411.04905} {Opencoder: The open cookbook for top-tier code large language models}.

\bibitem[{Islam et~al.(2023)Islam, Kannappan, Kiela, Qian, Scherrer, and Vidgen}]{islam2023financebench}
Pranab Islam, Anand Kannappan, Douwe Kiela, Rebecca Qian, Nino Scherrer, and Bertie Vidgen. 2023.
\newblock \href {https://arxiv.org/abs/2311.11944} {Financebench: A new benchmark for financial question answering}.
\newblock \emph{Preprint}, arXiv:2311.11944.

\bibitem[{Ivison et~al.(2024)Ivison, Wang, Liu, Wu, Pyatkin, Lambert, Smith, Choi, and Hajishirzi}]{ivison2024unpackingdpoppodisentangling}
Hamish Ivison, Yizhong Wang, Jiacheng Liu, Zeqiu Wu, Valentina Pyatkin, Nathan Lambert, Noah~A. Smith, Yejin Choi, and Hannaneh Hajishirzi. 2024.
\newblock \href {https://arxiv.org/abs/2406.09279} {Unpacking dpo and ppo: Disentangling best practices for learning from preference feedback}.

\bibitem[{Jiang et~al.(2023)Jiang, Sablayrolles, Mensch, Bamford, Chaplot, Casas, Bressand, Lengyel, Lample, Saulnier et~al.}]{jiang2023mistral}
Albert~Q Jiang, Alexandre Sablayrolles, Arthur Mensch, Chris Bamford, Devendra~Singh Chaplot, Diego de~las Casas, Florian Bressand, Gianna Lengyel, Guillaume Lample, Lucile Saulnier, et~al. 2023.
\newblock Mistral 7b.
\newblock \emph{arXiv preprint arXiv:2310.06825}.

\bibitem[{Jiao et~al.(2024)Jiao, Qin, Liu, Chen, and Joty}]{jiao-etal-2024-learning}
Fangkai Jiao, Chengwei Qin, Zhengyuan Liu, Nancy~F. Chen, and Shafiq Joty. 2024.
\newblock \href {https://aclanthology.org/2024.emnlp-main.20} {Learning planning-based reasoning by trajectories collection and process reward synthesizing}.
\newblock In \emph{Proceedings of the 2024 Conference on Empirical Methods in Natural Language Processing}, Miami, Florida, USA. Association for Computational Linguistics.

\bibitem[{Ke et~al.(2024)Ke, Kong, Li, Zhang, Mei, and Bendersky}]{bridging_retriever_llm_ke2024}
Zixuan Ke, Weize Kong, Cheng Li, Mingyang Zhang, Qiaozhu Mei, and Michael Bendersky. 2024.
\newblock Bridging the preference gap between retrievers and llms.
\newblock \emph{arXiv preprint arXiv:2401.06954}.

\bibitem[{Ke et~al.(2023)Ke, Shao, Lin, Konishi, Kim, and Liu}]{ke2023dgs}
Zixuan Ke, Yijia Shao, Haowei Lin, Tatsuya Konishi, Gyuhak Kim, and Bing Liu. 2023.
\newblock Continual pre-training of language models.
\newblock In \emph{International Conference on Learning Representations (ICLR)}.

\bibitem[{Ke et~al.(2022)Ke, Shao, Lin, Xu, Shu, and Liu}]{ke2022dga}
Zixuan Ke, Yijia Shao, Haowei Lin, Hu~Xu, Lei Shu, and Bing Liu. 2022.
\newblock Adapting a language model while preserving its general knowledge.
\newblock In \emph{Empirical Methods in Natural Language Processing (EMNLP)}.

\bibitem[{Khamnuansin et~al.(2024)Khamnuansin, Petchsod, Lertpiya, Balee, Lodkaew, Chalothorn, Pongthawornkamol, and Lertsutthiwong}]{labs2024thalletexthyperlocallyaugmented}
Danupat Khamnuansin, Atthakorn Petchsod, Anuruth Lertpiya, Pornchanan Balee, Thanawat Lodkaew, Tawunrat Chalothorn, Thadpong Pongthawornkamol, and Monchai Lertsutthiwong. 2024.
\newblock \href {https://arxiv.org/abs/2406.07505} {Thalle: Text hyperlocally augmented large language extension -- technical report}.
\newblock \emph{Preprint}, arXiv:2406.07505.

\bibitem[{Khot et~al.(2020)Khot, Clark, Guerquin, Jansen, and Sabharwal}]{allenai:qasc}
Tushar Khot, Peter Clark, Michal Guerquin, Peter Jansen, and Ashish Sabharwal. 2020.
\newblock Qasc: A dataset for question answering via sentence composition.
\newblock \emph{arXiv:1910.11473v2}.

\bibitem[{Kim et~al.(2022)Kim, Hessel, Jiang, West, Lu, Yu, Zhou, Bras, Alikhani, Kim, Sap, and Choi}]{kim2022soda}
Hyunwoo Kim, Jack Hessel, Liwei Jiang, Peter West, Ximing Lu, Youngjae Yu, Pei Zhou, Ronan~Le Bras, Malihe Alikhani, Gunhee Kim, Maarten Sap, and Yejin Choi. 2022.
\newblock Soda: Million-scale dialogue distillation with social commonsense contextualization.
\newblock \emph{ArXiv}, abs/2212.10465.

\bibitem[{Kwiatkowski et~al.(2019)Kwiatkowski, Palomaki, Redfield, Collins, Parikh, Alberti, Epstein, Polosukhin, Kelcey, Devlin, Lee, Toutanova, Jones, Chang, Dai, Uszkoreit, Le, and Petrov}]{nq-open}
Tom Kwiatkowski, Jennimaria Palomaki, Olivia Redfield, Michael Collins, Ankur Parikh, Chris Alberti, Danielle Epstein, Illia Polosukhin, Matthew Kelcey, Jacob Devlin, Kenton Lee, Kristina~N. Toutanova, Llion Jones, Ming-Wei Chang, Andrew Dai, Jakob Uszkoreit, Quoc Le, and Slav Petrov. 2019.
\newblock Natural questions: a benchmark for question answering research.
\newblock \emph{Transactions of the Association of Computational Linguistics}.

\bibitem[{Lambert et~al.(2024)Lambert, Morrison, Pyatkin, Huang, Ivison, Brahman, Miranda, Liu, Dziri, Lyu, Gu, Malik, Graf, Hwang, Yang, Bras, Tafjord, Wilhelm, Soldaini, Smith, Wang, Dasigi, and Hajishirzi}]{lambert2024tulu3}
Nathan Lambert, Jacob Morrison, Valentina Pyatkin, Shengyi Huang, Hamish Ivison, Faeze Brahman, Lester James~V. Miranda, Alisa Liu, Nouha Dziri, Shane Lyu, Yuling Gu, Saumya Malik, Victoria Graf, Jena~D. Hwang, Jiangjiang Yang, Ronan~Le Bras, Oyvind Tafjord, Chris Wilhelm, Luca Soldaini, Noah~A. Smith, Yizhong Wang, Pradeep Dasigi, and Hannaneh Hajishirzi. 2024.
\newblock Tülu 3: Pushing frontiers in open language model post-training.

\bibitem[{Lewis et~al.(2020)Lewis, Perez, Piktus, Petroni, Karpukhin, Goyal, K{\"u}ttler, Lewis, Yih, Rockt{\"a}schel et~al.}]{lewis2020retrieval}
Patrick Lewis, Ethan Perez, Aleksandra Piktus, Fabio Petroni, Vladimir Karpukhin, Naman Goyal, Heinrich K{\"u}ttler, Mike Lewis, Wen-tau Yih, Tim Rockt{\"a}schel, et~al. 2020.
\newblock Retrieval-augmented generation for knowledge-intensive nlp tasks.
\newblock \emph{Advances in Neural Information Processing Systems}, 33:9459--9474.

\bibitem[{Li et~al.(2023)Li, Wang, Ding, and Chen}]{lwdc23}
Yinheng Li, Shaofei Wang, Han Ding, and Hang Chen. 2023.
\newblock Large language models in finance: A survey.
\newblock In \emph{Proceedings of the Fourth ACM International Conference on AI in Finance}, pages 374--382.

\bibitem[{Lian et~al.(2023)Lian, Wang, Goodson, Pentland, Cook, Vong, and "Teknium"}]{SlimOrca}
Wing Lian, Guan Wang, Bleys Goodson, Eugene Pentland, Austin Cook, Chanvichet Vong, and "Teknium". 2023.
\newblock \href {https://https://huggingface.co/Open-Orca/SlimOrca} {Slimorca: An open dataset of gpt-4 augmented flan reasoning traces, with verification}.

\bibitem[{Lightman et~al.(2024)Lightman, Kosaraju, Burda, Edwards, Baker, Lee, Leike, Schulman, Sutskever, and Cobbe}]{lightman2024lets}
Hunter Lightman, Vineet Kosaraju, Yuri Burda, Harrison Edwards, Bowen Baker, Teddy Lee, Jan Leike, John Schulman, Ilya Sutskever, and Karl Cobbe. 2024.
\newblock \href {https://openreview.net/forum?id=v8L0pN6EOi} {Let's verify step by step}.
\newblock In \emph{The Twelfth International Conference on Learning Representations}.

\bibitem[{Ling et~al.(2017)Ling, Yogatama, Dyer, and Blunsom}]{ling2017program}
Wang Ling, Dani Yogatama, Chris Dyer, and Phil Blunsom. 2017.
\newblock Program induction by rationale generation: Learning to solve and explain algebraic word problems.
\newblock \emph{ACL}.

\bibitem[{Liu et~al.(2024)Liu, Geng, Wu, Sucholutsky, Lombrozo, and Griffiths}]{liu2024mindstepbystep}
Ryan Liu, Jiayi Geng, Addison~J. Wu, Ilia Sucholutsky, Tania Lombrozo, and Thomas~L. Griffiths. 2024.
\newblock \href {https://arxiv.org/abs/2410.21333} {Mind your step (by step): Chain-of-thought can reduce performance on tasks where thinking makes humans worse}.
\newblock \emph{Preprint}, arXiv:2410.21333.

\bibitem[{LLaMA(2024)}]{grattafiori2024llama3herdmodels}
Team LLaMA. 2024.
\newblock \href {https://arxiv.org/abs/2407.21783} {The llama 3 herd of models}.
\newblock \emph{Preprint}, arXiv:2407.21783.

\bibitem[{Luo et~al.(2024)Luo, Liu, Liu, Phatale, Lara, Li, Shu, Zhu, Meng, Sun, and Rastogi}]{luo2024improvemathematicalreasoninglanguage}
Liangchen Luo, Yinxiao Liu, Rosanne Liu, Samrat Phatale, Harsh Lara, Yunxuan Li, Lei Shu, Yun Zhu, Lei Meng, Jiao Sun, and Abhinav Rastogi. 2024.
\newblock \href {https://arxiv.org/abs/2406.06592} {Improve mathematical reasoning in language models by automated process supervision}.
\newblock \emph{Preprint}, arXiv:2406.06592.

\bibitem[{Luo et~al.(2023)Luo, Xu, Zhao, Sun, Geng, Hu, Tao, Ma, Lin, and Jiang}]{luo2023wizardcoder}
Ziyang Luo, Can Xu, Pu~Zhao, Qingfeng Sun, Xiubo Geng, Wenxiang Hu, Chongyang Tao, Jing Ma, Qingwei Lin, and Daxin Jiang. 2023.
\newblock Wizardcoder: Empowering code large language models with evol-instruct.

\bibitem[{Madaan et~al.(2023)Madaan, Tandon, Gupta, Hallinan, Gao, Wiegreffe, Alon, Dziri, Prabhumoye, Yang, Gupta, Majumder, Hermann, Welleck, Yazdanbakhsh, and Clark}]{madaan2023selfrefine}
Aman Madaan, Niket Tandon, Prakhar Gupta, Skyler Hallinan, Luyu Gao, Sarah Wiegreffe, Uri Alon, Nouha Dziri, Shrimai Prabhumoye, Yiming Yang, Shashank Gupta, Bodhisattwa~Prasad Majumder, Katherine Hermann, Sean Welleck, Amir Yazdanbakhsh, and Peter Clark. 2023.
\newblock \href {https://openreview.net/forum?id=S37hOerQLB} {Self-refine: Iterative refinement with self-feedback}.
\newblock In \emph{Thirty-seventh Conference on Neural Information Processing Systems}.

\bibitem[{Maia et~al.(2018)Maia, Berant, Farinha, and Freitas}]{maia201818}
Mário Maia, Jonathan Berant, José Farinha, and André Freitas. 2018.
\newblock Www'18 open challenge: Financial opinion mining and question answering.
\newblock In \emph{Proceedings of the 2018 World Wide Web Conference}, pages 1941--1942. International World Wide Web Conferences Steering Committee.

\bibitem[{Malmgren(2014)}]{textract}
Dean Malmgren. 2014.
\newblock \href {https://textract.readthedocs.io/} {Textract}.

\bibitem[{Malo et~al.(2014)Malo, Sinha, Korhonen, Wallenius, and Takala}]{Malo2014GoodDO}
P.~Malo, A.~Sinha, P.~Korhonen, J.~Wallenius, and P.~Takala. 2014.
\newblock Good debt or bad debt: Detecting semantic orientations in economic texts.
\newblock \emph{Journal of the Association for Information Science and Technology}, 65.

\bibitem[{Mihaylov et~al.(2018)Mihaylov, Clark, Khot, and Sabharwal}]{OpenBookQA2018}
Todor Mihaylov, Peter Clark, Tushar Khot, and Ashish Sabharwal. 2018.
\newblock Can a suit of armor conduct electricity? a new dataset for open book question answering.
\newblock In \emph{EMNLP}.

\bibitem[{Mishra et~al.(2022)Mishra, Khashabi, Baral, and Hajishirzi}]{naturalinstructions}
Swaroop Mishra, Daniel Khashabi, Chitta Baral, and Hannaneh Hajishirzi. 2022.
\newblock Cross-task generalization via natural language crowdsourcing instructions.
\newblock In \emph{ACL}.

\bibitem[{Mitra et~al.(2024)Mitra, Khanpour, Rosset, and Awadallah}]{mitra2024orcamath}
Arindam Mitra, Hamed Khanpour, Corby Rosset, and Ahmed Awadallah. 2024.
\newblock \href {https://arxiv.org/abs/2402.14830} {Orca-math: Unlocking the potential of slms in grade school math}.
\newblock \emph{Preprint}, arXiv:2402.14830.

\bibitem[{Mukherjee et~al.(2022)Mukherjee, Bohra, Banerjee, Sharma, Hegde, Shaikh, Shrivastava, Dasgupta, Ganguly, Ghosh, and Goyal}]{mukherjee-etal-2022-ectsum}
Rajdeep Mukherjee, Abhinav Bohra, Akash Banerjee, Soumya Sharma, Manjunath Hegde, Afreen Shaikh, Shivani Shrivastava, Koustuv Dasgupta, Niloy Ganguly, Saptarshi Ghosh, and Pawan Goyal. 2022.
\newblock \href {https://aclanthology.org/2022.emnlp-main.748} {{ECTS}um: A new benchmark dataset for bullet point summarization of long earnings call transcripts}.
\newblock In \emph{Proceedings of the 2022 Conference on Empirical Methods in Natural Language Processing}, Abu Dhabi, United Arab Emirates. Association for Computational Linguistics.

\bibitem[{Onoe et~al.(2021)Onoe, Zhang, Choi, and Durrett}]{onoe2021creak}
Yasumasa Onoe, Michael~J.Q. Zhang, Eunsol Choi, and Greg Durrett. 2021.
\newblock Creak: A dataset for commonsense reasoning over entity knowledge.
\newblock \emph{OpenReview}.

\bibitem[{OpenAI(2023)}]{aaa+23}
OpenAI. 2023.
\newblock Gpt-4 technical report.
\newblock \emph{arXiv preprint arXiv:2303.08774}.

\bibitem[{Pang et~al.(2024)Pang, Yuan, He, Cho, Sukhbaatar, and Weston}]{pang2024iterative}
Richard~Yuanzhe Pang, Weizhe Yuan, He~He, Kyunghyun Cho, Sainbayar Sukhbaatar, and Jason~E Weston. 2024.
\newblock \href {https://openreview.net/forum?id=4XIKfvNYvx} {Iterative reasoning preference optimization}.
\newblock In \emph{The Thirty-eighth Annual Conference on Neural Information Processing Systems}.

\bibitem[{Penedo et~al.(2024)Penedo, Kydlíček, allal, Lozhkov, Mitchell, Raffel, Werra, and Wolf}]{penedo2024finewebdatasetsdecantingweb}
Guilherme Penedo, Hynek Kydlíček, Loubna~Ben allal, Anton Lozhkov, Margaret Mitchell, Colin Raffel, Leandro~Von Werra, and Thomas Wolf. 2024.
\newblock \href {https://arxiv.org/abs/2406.17557} {The fineweb datasets: Decanting the web for the finest text data at scale}.
\newblock \emph{Preprint}, arXiv:2406.17557.

\bibitem[{Quinlan(1987)}]{statlog_(australian_credit_approval)_143}
Ross Quinlan. 1987.
\newblock {Statlog (Australian Credit Approval)}.
\newblock UCI Machine Learning Repository.
\newblock {DOI}: https://doi.org/10.24432/C59012.

\bibitem[{Rafailov et~al.(2023)Rafailov, Sharma, Mitchell, Manning, Ermon, and Finn}]{rafailov2023direct}
Rafael Rafailov, Archit Sharma, Eric Mitchell, Christopher~D Manning, Stefano Ermon, and Chelsea Finn. 2023.
\newblock \href {https://openreview.net/forum?id=HPuSIXJaa9} {Direct preference optimization: Your language model is secretly a reward model}.
\newblock In \emph{Thirty-seventh Conference on Neural Information Processing Systems}.

\bibitem[{Raffel et~al.(2020)Raffel, Shazeer, Roberts, Lee, Narang, Matena, Zhou, Li, and Liu}]{2020t5}
Colin Raffel, Noam Shazeer, Adam Roberts, Katherine Lee, Sharan Narang, Michael Matena, Yanqi Zhou, Wei Li, and Peter~J. Liu. 2020.
\newblock \href {http://jmlr.org/papers/v21/20-074.html} {Exploring the limits of transfer learning with a unified text-to-text transformer}.
\newblock \emph{Journal of Machine Learning Research}, 21(140):1--67.

\bibitem[{Sakaguchi et~al.(2019)Sakaguchi, Bras, Bhagavatula, and Choi}]{sakaguchi2019winogrande}
Keisuke Sakaguchi, Ronan~Le Bras, Chandra Bhagavatula, and Yejin Choi. 2019.
\newblock Winogrande: An adversarial winograd schema challenge at scale.
\newblock \emph{arXiv preprint arXiv:1907.10641}.

\bibitem[{Saunders et~al.(2022)Saunders, Yeh, Wu, Bills, Ouyang, Ward, and Leike}]{saunders2022selfcritiquingmodelsassistinghuman}
William Saunders, Catherine Yeh, Jeff Wu, Steven Bills, Long Ouyang, Jonathan Ward, and Jan Leike. 2022.
\newblock \href {https://arxiv.org/abs/2206.05802} {Self-critiquing models for assisting human evaluators}.
\newblock \emph{Preprint}, arXiv:2206.05802.

\bibitem[{Saxton et~al.(2019)Saxton, Grefenstette, Hill, and Kohli}]{2019arXiv}
Saxton, Grefenstette, Hill, and Kohli. 2019.
\newblock Analysing mathematical reasoning abilities of neural models.
\newblock \emph{arXiv:1904.01557}.

\bibitem[{Scialom et~al.(2022)Scialom, Chakrabarty, and Muresan}]{scialom-etal-2022-fine}
Thomas Scialom, Tuhin Chakrabarty, and Smaranda Muresan. 2022.
\newblock \href {https://doi.org/10.18653/v1/2022.emnlp-main.410} {Fine-tuned language models are continual learners}.
\newblock In \emph{Proceedings of the 2022 Conference on Empirical Methods in Natural Language Processing}, Abu Dhabi, United Arab Emirates. Association for Computational Linguistics.

\bibitem[{Setlur et~al.(2024)Setlur, Nagpal, Fisch, Geng, Eisenstein, Agarwal, Agarwal, Berant, and Kumar}]{setlur2024rewardingprogressscalingautomated}
Amrith Setlur, Chirag Nagpal, Adam Fisch, Xinyang Geng, Jacob Eisenstein, Rishabh Agarwal, Alekh Agarwal, Jonathan Berant, and Aviral Kumar. 2024.
\newblock \href {https://arxiv.org/abs/2410.08146} {Rewarding progress: Scaling automated process verifiers for llm reasoning}.
\newblock \emph{Preprint}, arXiv:2410.08146.

\bibitem[{Shah et~al.(2023)Shah, Paturi, and Chava}]{shah-etal-2023-trillion}
Agam Shah, Suvan Paturi, and Sudheer Chava. 2023.
\newblock \href {https://aclanthology.org/2023.acl-long.368} {Trillion dollar words: A new financial dataset, task {\&} market analysis}.
\newblock In \emph{Proceedings of the 61st Annual Meeting of the Association for Computational Linguistics (Volume 1: Long Papers)}, Toronto, Canada. Association for Computational Linguistics.

\bibitem[{Sharma et~al.(2022)Sharma, Nayak, Bose, Meena, Dasgupta, Ganguly, and Goyal}]{sharma2022finred}
Soumya Sharma, Tapas Nayak, Arusarka Bose, Ajay~Kumar Meena, Koustuv Dasgupta, Niloy Ganguly, and Pawan Goyal. 2022.
\newblock Finred: A dataset for relation extraction in financial domain.
\newblock In \emph{Proceedings of The 2nd Workshop on Financial Technology on the Web (FinWeb)}.

\bibitem[{Sinha et~al.(2020)Sinha, Khandait, and vincent}]{DBLP:journals/corr/abs-2009-04202}
Ankur Sinha, Tanmay Khandait, and vincent. 2020.
\newblock \href {https://arxiv.org/abs/2009.04202} {Impact of news on the commodity market: Dataset and results}.
\newblock \emph{CoRR}, abs/2009.04202.

\bibitem[{Snell et~al.(2024)Snell, Lee, Xu, and Kumar}]{snell2024scalingllmtesttimecompute}
Charlie Snell, Jaehoon Lee, Kelvin Xu, and Aviral Kumar. 2024.
\newblock \href {https://arxiv.org/abs/2408.03314} {Scaling llm test-time compute optimally can be more effective than scaling model parameters}.
\newblock \emph{Preprint}, arXiv:2408.03314.

\bibitem[{Soun et~al.(2022)Soun, Yoo, Cho, Jeon, and Kang}]{soun2022accurate}
Yejun Soun, Jaemin Yoo, Minyong Cho, Jihyeong Jeon, and U~Kang. 2022.
\newblock Accurate stock movement prediction with self-supervised learning from sparse noisy tweets.
\newblock In \emph{2022 IEEE International Conference on Big Data (Big Data)}, pages 1691--1700. IEEE Computer Society.

\bibitem[{Sprague et~al.(2024)Sprague, Yin, Rodriguez, Jiang, Wadhwa, Singhal, Zhao, Ye, Mahowald, and Durrett}]{sprague2024cotcotchainofthoughthelps}
Zayne Sprague, Fangcong Yin, Juan~Diego Rodriguez, Dongwei Jiang, Manya Wadhwa, Prasann Singhal, Xinyu Zhao, Xi~Ye, Kyle Mahowald, and Greg Durrett. 2024.
\newblock \href {https://arxiv.org/abs/2409.12183} {To cot or not to cot? chain-of-thought helps mainly on math and symbolic reasoning}.
\newblock \emph{Preprint}, arXiv:2409.12183.

\bibitem[{Tunstall et~al.(2024)Tunstall, Beeching, Lambert, Rajani, Rasul, Belkada, Huang, Werra, Fourrier, Habib, Sarrazin, Sanseviero, Rush, and Wolf}]{tunstall2024zephyr}
Lewis Tunstall, Edward~Emanuel Beeching, Nathan Lambert, Nazneen Rajani, Kashif Rasul, Younes Belkada, Shengyi Huang, Leandro~Von Werra, Cl{\'e}mentine Fourrier, Nathan Habib, Nathan Sarrazin, Omar Sanseviero, Alexander~M Rush, and Thomas Wolf. 2024.
\newblock \href {https://openreview.net/forum?id=aKkAwZB6JV} {Zephyr: Direct distillation of {LM} alignment}.
\newblock In \emph{First Conference on Language Modeling}.

\bibitem[{Wang et~al.(2022)Wang, Kordi, Mishra, Liu, Smith, Khashabi, and Hajishirzi}]{selfinstruct}
Yizhong Wang, Yeganeh Kordi, Swaroop Mishra, Alisa Liu, Noah~A. Smith, Daniel Khashabi, and Hannaneh Hajishirzi. 2022.
\newblock Self-instruct: Aligning language model with self generated instructions.

\bibitem[{Wei et~al.(2022)Wei, Bosma, Zhao, Guu, Yu, Lester, Du, Dai, and Le}]{wei2022finetuned}
Jason Wei, Maarten Bosma, Vincent Zhao, Kelvin Guu, Adams~Wei Yu, Brian Lester, Nan Du, Andrew~M. Dai, and Quoc~V Le. 2022.
\newblock \href {https://openreview.net/forum?id=gEZrGCozdqR} {Finetuned language models are zero-shot learners}.
\newblock In \emph{International Conference on Learning Representations}.

\bibitem[{Wei et~al.(2023)Wei, Wang, Schuurmans, Bosma, Ichter, Xia, Chi, Le, and Zhou}]{wei2023chainofthoughtpromptingelicitsreasoning}
Jason Wei, Xuezhi Wang, Dale Schuurmans, Maarten Bosma, Brian Ichter, Fei Xia, Ed~Chi, Quoc Le, and Denny Zhou. 2023.
\newblock \href {https://arxiv.org/abs/2201.11903} {Chain-of-thought prompting elicits reasoning in large language models}.
\newblock \emph{Preprint}, arXiv:2201.11903.

\bibitem[{Welbl et~al.(2017)Welbl, Liu, and Gardner}]{Welbl2017CrowdsourcingMC}
Johannes Welbl, Nelson~F. Liu, and Matt Gardner. 2017.
\newblock Crowdsourcing multiple choice science questions.
\newblock In \emph{NUT@EMNLP}.

\bibitem[{Writer(2024)}]{Palmyra-Fin-70B-32k}
Writer. 2024.
\newblock {Palmyra-Fin-70B-32k: a powerful LLM designed for Finance}.
\newblock \url{https://dev.writer.com}.

\bibitem[{Wu et~al.(2018)Wu, Zhang, Shen, and Wang}]{WuCIKM18}
Huizhe Wu, Wei Zhang, Weiwei Shen, and Jun Wang. 2018.
\newblock \href {https://doi.org/10.1145/3269206.3269290} {Hybrid deep sequential modeling for social text-driven stock prediction}.
\newblock In \emph{Proceedings of the 27th ACM International Conference on Information and Knowledge Management}, CIKM '18, page 1627–1630, New York, NY, USA. Association for Computing Machinery.

\bibitem[{Xie et~al.(2023)Xie, Han, Zhang, Lai, Peng, Lopez-Lira, and Huang}]{xie2023pixiulargelanguagemodel}
Qianqian Xie, Weiguang Han, Xiao Zhang, Yanzhao Lai, Min Peng, Alejandro Lopez-Lira, and Jimin Huang. 2023.
\newblock \href {https://arxiv.org/abs/2306.05443} {Pixiu: A large language model, instruction data and evaluation benchmark for finance}.
\newblock \emph{Preprint}, arXiv:2306.05443.

\bibitem[{Xie et~al.(2024{\natexlab{a}})Xie, Li, Xiao, Jiang, Xiang, Zhang, Chen, He, Han, Yang, Chen, Zhang, Shen, Kim, Liu, Luo, Yu, Cao, Deng, Yao, Li, Feng, Dai, Somasundaram, Lu, Zhao, Long, Xiong, Smith, Yu, Lai, Peng, Nie, Suchow, Liu, Wang, Lopez-Lira, Huang, and Ananiadou}]{xie2024openfinllmsopenmultimodallarge}
Qianqian Xie, Dong Li, Mengxi Xiao, Zihao Jiang, Ruoyu Xiang, Xiao Zhang, Zhengyu Chen, Yueru He, Weiguang Han, Yuzhe Yang, Shunian Chen, Yifei Zhang, Lihang Shen, Daniel Kim, Zhiwei Liu, Zheheng Luo, Yangyang Yu, Yupeng Cao, Zhiyang Deng, Zhiyuan Yao, Haohang Li, Duanyu Feng, Yongfu Dai, VijayaSai Somasundaram, Peng Lu, Yilun Zhao, Yitao Long, Guojun Xiong, Kaleb Smith, Honghai Yu, Yanzhao Lai, Min Peng, Jianyun Nie, Jordan~W. Suchow, Xiao-Yang Liu, Benyou Wang, Alejandro Lopez-Lira, Jimin Huang, and Sophia Ananiadou. 2024{\natexlab{a}}.
\newblock \href {https://arxiv.org/abs/2408.11878} {Open-finllms: Open multimodal large language models for financial applications}.
\newblock \emph{Preprint}, arXiv:2408.11878.

\bibitem[{Xie et~al.(2022)Xie, Wu, Shi, Zhong, Scholak, Yasunaga, Wu, Zhong, Yin, Wang, Zhong, Wang, Li, Boyle, Ni, Yao, Radev, Xiong, Kong, Zhang, Smith, Zettlemoyer, and Yu}]{UnifiedSKG}
Tianbao Xie, Chen~Henry Wu, Peng Shi, Ruiqi Zhong, Torsten Scholak, Michihiro Yasunaga, Chien-Sheng Wu, Ming Zhong, Pengcheng Yin, Sida~I. Wang, Victor Zhong, Bailin Wang, Chengzu Li, Connor Boyle, Ansong Ni, Ziyu Yao, Dragomir Radev, Caiming Xiong, Lingpeng Kong, Rui Zhang, Noah~A. Smith, Luke Zettlemoyer, and Tao Yu. 2022.
\newblock Unifiedskg: Unifying and multi-tasking structured knowledge grounding with text-to-text language models.
\newblock \emph{EMNLP}.

\bibitem[{Xie et~al.(2024{\natexlab{b}})Xie, Aggarwal, and Ahmad}]{xie-etal-2024-efficient}
Yong Xie, Karan Aggarwal, and Aitzaz Ahmad. 2024{\natexlab{b}}.
\newblock \href {https://doi.org/10.18653/v1/2024.findings-acl.606} {Efficient continual pre-training for building domain specific large language models}.
\newblock In \emph{Findings of the Association for Computational Linguistics: ACL 2024}, pages 10184--10201, Bangkok, Thailand. Association for Computational Linguistics.

\bibitem[{Xu and Cohen(2018)}]{xu-cohen-2018-stock}
Yumo Xu and Shay~B. Cohen. 2018.
\newblock \href {https://aclanthology.org/P18-1183} {Stock movement prediction from tweets and historical prices}.
\newblock In \emph{Proceedings of the 56th Annual Meeting of the Association for Computational Linguistics (Volume 1: Long Papers)}, Melbourne, Australia. Association for Computational Linguistics.

\bibitem[{Yang et~al.(2023)Yang, Liu, and Wang}]{yang2023fingpt}
Hongyang Yang, Xiao-Yang Liu, and Christina~Dan Wang. 2023.
\newblock Fingpt: Open-source financial large language models.
\newblock \emph{FinLLM Symposium at IJCAI 2023}.

\bibitem[{Yang et~al.(2020)Yang, Kenny, Ng, Yang, Smyth, and Dong}]{Yang2020GeneratingPC}
Linyi Yang, Eoin~M. Kenny, Tin Lok~James Ng, K.~Z. Zhang P.K. Kannan~Y Yang, Barry Smyth, and Ruihai Dong. 2020.
\newblock Generating plausible counterfactual explanations for deep transformers in financial text classification.
\newblock In \emph{COLING}.

\bibitem[{Yu et~al.(2023)Yu, Jiang, Shi, Yu, Liu, Zhang, Kwok, Li, Weller, and Liu}]{yu2023metamath}
Longhui Yu, Weisen Jiang, Han Shi, Jincheng Yu, Zhengying Liu, Yu~Zhang, James~T Kwok, Zhenguo Li, Adrian Weller, and Weiyang Liu. 2023.
\newblock Metamath: Bootstrap your own mathematical questions for large language models.
\newblock \emph{arXiv preprint arXiv:2309.12284}.

\bibitem[{Yuan et~al.(2024)Yuan, Pang, Cho, Li, Sukhbaatar, Xu, and Weston}]{yuan2024selfrewarding}
Weizhe Yuan, Richard~Yuanzhe Pang, Kyunghyun Cho, Xian Li, Sainbayar Sukhbaatar, Jing Xu, and Jason~E Weston. 2024.
\newblock \href {https://openreview.net/forum?id=0NphYCmgua} {Self-rewarding language models}.
\newblock In \emph{Forty-first International Conference on Machine Learning}.

\bibitem[{Yue et~al.(2023)Yue, Qu, Zhang, Fu, Huang, Sun, Su, and Chen}]{yue2023mammoth}
Xiang Yue, Xingwei Qu, Ge~Zhang, Yao Fu, Wenhao Huang, Huan Sun, Yu~Su, and Wenhu Chen. 2023.
\newblock Mammoth: Building math generalist models through hybrid instruction tuning.
\newblock \emph{arXiv preprint arXiv:2309.05653}.

\bibitem[{Zellers et~al.(2019)Zellers, Holtzman, Bisk, Farhadi, and Choi}]{zellers2019hellaswag}
Rowan Zellers, Ari Holtzman, Yonatan Bisk, Ali Farhadi, and Yejin Choi. 2019.
\newblock Hellaswag: Can a machine really finish your sentence?
\newblock In \emph{Proceedings of the 57th Annual Meeting of the Association for Computational Linguistics}.

\bibitem[{Zhang et~al.(2023)Zhang, Qian, Liu, Heinecke, Meng, Liu, Yu, Savarese, and Xiong}]{zhang2023dialogstudio}
Jianguo Zhang, Kun Qian, Zhiwei Liu, Shelby Heinecke, Rui Meng, Ye~Liu, Zhou Yu, Silvio Savarese, and Caiming Xiong. 2023.
\newblock Dialogstudio: Towards richest and most diverse unified dataset collection for conversational ai.
\newblock \emph{arXiv preprint arXiv:2307.10172}.

\bibitem[{Zheng et~al.(2023)Zheng, Chiang, Sheng, Zhuang, Wu, Zhuang, Lin, Li, Li, Xing, Zhang, Gonzalez, and Stoica}]{zheng2023judging}
Lianmin Zheng, Wei-Lin Chiang, Ying Sheng, Siyuan Zhuang, Zhanghao Wu, Yonghao Zhuang, Zi~Lin, Zhuohan Li, Dacheng Li, Eric.~P Xing, Hao Zhang, Joseph~E. Gonzalez, and Ion Stoica. 2023.
\newblock \href {https://arxiv.org/abs/2306.05685} {Judging llm-as-a-judge with mt-bench and chatbot arena}.
\newblock \emph{Preprint}, arXiv:2306.05685.

\bibitem[{Zhou et~al.(2021)Zhou, Ma, and Liu}]{zhou-etal-2021-trade}
Zhihan Zhou, Liqian Ma, and Han Liu. 2021.
\newblock \href {https://doi.org/10.18653/v1/2021.findings-acl.186} {Trade the event: Corporate events detection for news-based event-driven trading}.
\newblock In \emph{Findings of the Association for Computational Linguistics: ACL-IJCNLP 2021}, pages 2114--2124, Online. Association for Computational Linguistics.

\bibitem[{Zhu et~al.(2021)Zhu, Lei, Huang, Wang, Zhang, Lv, Feng, and Chua}]{zhu-etal-2021-tat}
Fengbin Zhu, Wenqiang Lei, Youcheng Huang, Chao Wang, Shuo Zhang, Jiancheng Lv, Fuli Feng, and Tat-Seng Chua. 2021.
\newblock \href {https://aclanthology.org/2021.acl-long.254} {{TAT}-{QA}: A question answering benchmark on a hybrid of tabular and textual content in finance}.
\newblock In \emph{Proceedings of the 59th Annual Meeting of the Association for Computational Linguistics and the 11th International Joint Conference on Natural Language Processing (Volume 1: Long Papers)}, Online. Association for Computational Linguistics.

\end{thebibliography}
\appendix

\newpage
\counterwithin{table}{section}
\renewcommand{\thetable}{\Alph{section}.\arabic{table}}
\counterwithin{figure}{section}
\renewcommand{\thefigure}{\Alph{section}.\arabic{figure}}

\section{Preventing Data Contamination}
\label{ap.contamination}

When designing FinEval, we took extra care to ensure that evaluation tasks do not duplicate any samples from FinTrain. To further verify this, we followed your suggestion and computed string matches between FinEval and FinTrain.

Specifically, we adopted the decontamination procedure described in the Hugging Face blog you referenced. A training sample is contaminated if it is overlapped with any evaluation sample. The contamination ratio is computed as the fraction of contamination samples in the training samples. Based on this criterion, we report two contamination ratios:

\begin{itemize}
    \item \textbf{0\%} under the strictest setting, where only complete sample matches are considered contamination.
    \item \textbf{0.003\%} using the method described in the blog—where 10-gram matches are used for pre-identification, followed by difflib.SequenceMatcher. If over 50\% of its characters match any of the evaluation samples, the training sample is considered contaminated.
\end{itemize}

These contamination rates are extremely low, indicating minimal overlap between FinTrain and FinEval. Upon manual inspection of the few samples flagged by the 50\% character overlap rule, we found they involve either (1) partial overlap in the question format or instruction prompt, which is expected for the similar tasks where the task type (e.g., sentiment analysis) has been seen, but the content remain unseen; or (2) partial overlap in the input content (e.g., shared elements in bank transcripts), but the specific question and answers are unseen. In both cases, these do not indicate memorization or leakage of benchmark content.


\section{Ablations and Understanding \ourframework}
\label{sec.ablation}




\subsection{Continual Pre-training}
\label{sec.cpt}

In order to expose the LLM to domain-specific concepts, we first conduct continual pre-training (CPT). In CPT, we feed plain text to the LLM and perform \textit{next token prediction}.

\noindent\textbf{From Text to CPT Data}. A key challenge in CPT is what kind of data we should use.
Given the general and domain-specific texts introduced in \S\ref{sec.text_curate}, we can construct three versions of CPT data, \textit{CPT-In} contains only the financial (in-domain) text, \textit{CPT-Gen} contains only the general domain data, and \textit{CPT-Mix} contains the mixture of the CPT-In and CPT-Gen. 

\begin{figure}[h]
\centering
\includegraphics[width=\columnwidth]{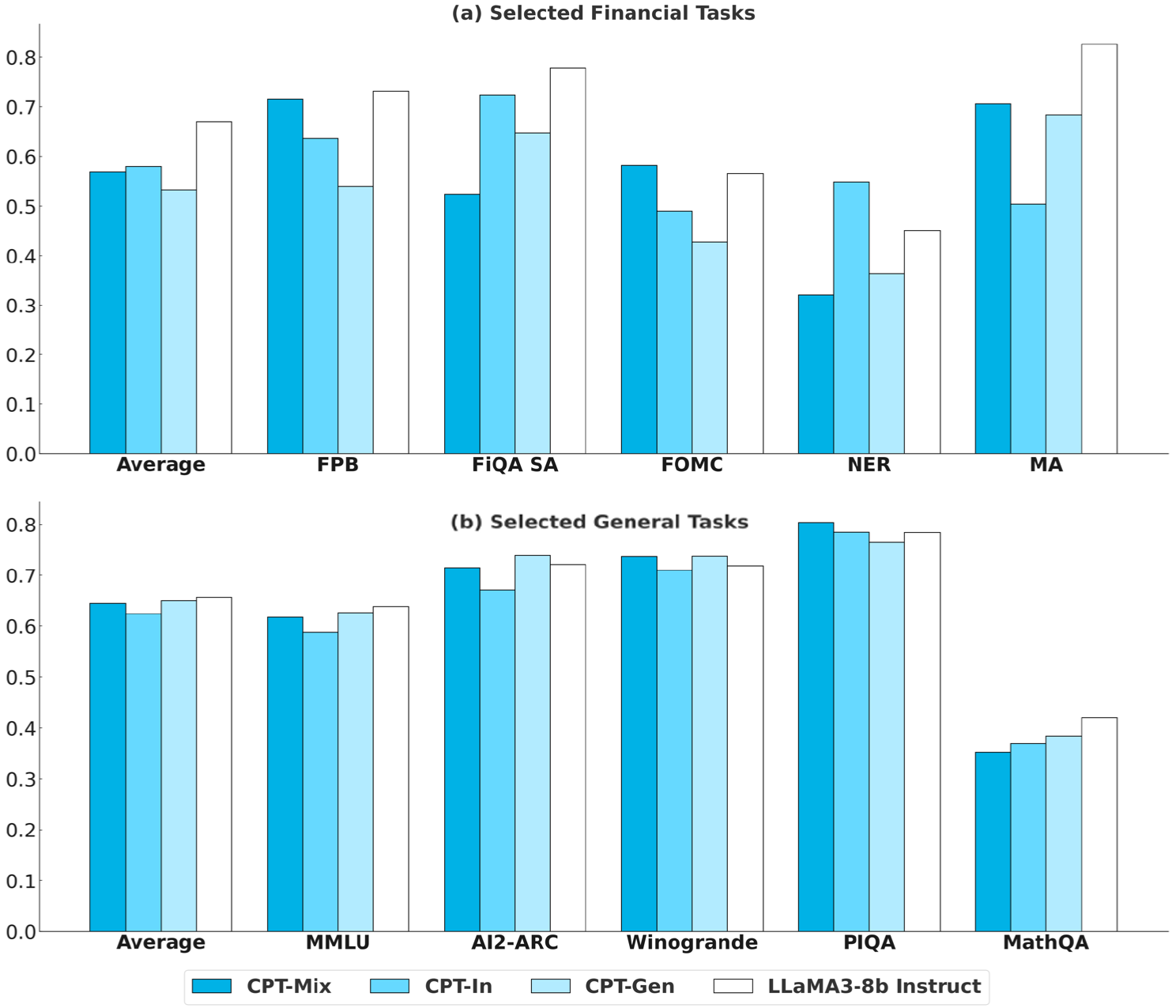}
\caption{Average performance on selected datasets for training Llama3-8b-instruct on our CPT-In, CPT-Gen and CPT-Mix. The Y-axis represents the same performance metrics as those reported in Tables~\ref{tab.similar_results} and \ref{tab.novel_results}. The selected datasets are chosen for illustration purpose based on their ability to illustrate the general trend.
}
\label{fig.cpt_combined}
\end{figure}

\noindent\textbf{Key Data Experiments.} We conduct CPT on each of the three versions of data. As shown in Figure~\ref{fig.cpt_combined}, we observe that while CPT-In and CPT-Gen outperforms in financial (Fig~\ref{fig.cpt_combined}a) and general (Fig~\ref{fig.cpt_combined}b) tasks, respectively, CPT-Mix achieves the best overall. This is expected as CPT-In can cause \textit{catastrophic forgetting} on the general tasks, while incorporating general domain concepts in CPT-Mix acts as `replay' mechanism to mitigate it \citep{scialom-etal-2022-fine}. 
We can also see that none of the CPT-trained LLMs outperform their base. This is unexpected because CPT invovles post-training on more specialized data, which should enhance the performance. By analyzing the output, we attribute this issue to the model forgetting how to follow instructions effectively after CPT. To quantify this finding, we evaluate the instruction following ability of these models using MT-Bench. The two-turn average scores for CPT-Mix, CPT-In, and CPT-Gen are 1, 1, and 1.0125, respectively, while the base model, achieves a score of 7.8875. These confirm that the conventional CPT applied to a instruction-tuned LLM can cause serious forgetting on instruction-following (IF) capability. In \S\ref{sec.it}, we will see how jointly train IT and CPT can help mitigate such forgetting issue.

\subsection{Instruction Following}
\label{sec.it}

To adapt the LLM to domain-specific and IF tasks,  we conduct IT. The key different between IT and CPT is that IT \textit{masks out the instruction} and \textit{takes as input supervised tasks}. 

\noindent\textbf{From Prompt to IT Data.} 
We introduced our prompt curation in \S\ref{sec.prompt_curate}. We create the responses for IT by \textit{filtering existing responses} or \textit{creating new responses}. For prompts with existing responses, we generally keep the original responses if they were written by a human or a strong model, such as GPT-4. We also filter out empty responses. For prompts without responses, for example, exercises extracted from books that may not have solutions provided, we generate new responses using GPT-4o. Similar to CPT data, we construct three versions of IT data, \textit{IT-In}, which contains only financial (in-domain) tasks, \textit{IT-Gen}, which contains only general tasks, and \textit{IT-Mix}, which includes a mixture of the IT-In and IT-Gen.

\begin{figure}[h]
\centering
\includegraphics[width=\columnwidth]{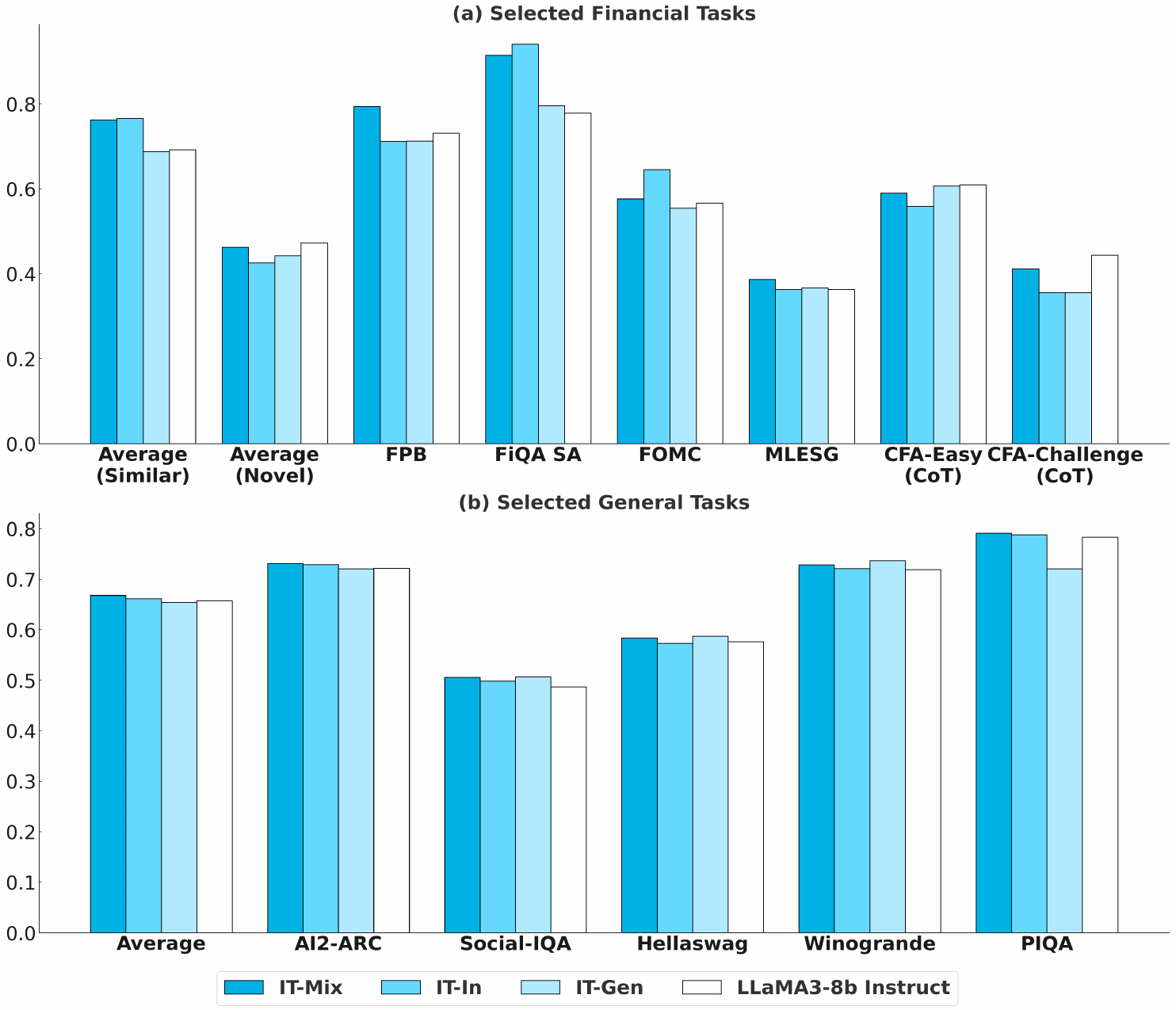}
\caption{Average performance on selected datasets for training Llama3-8b-instruct on our IT-In, IT-Gen and IT-Mix. 
}
\label{fig.it_combined}
\vspace{-2mm}
\end{figure}

\noindent\textbf{Key Data Experiments.} 
Similar to CPT, we conduct IT to each of the three versions. From Figure~\ref{fig.it_combined}, we observe that unlike CPT, forgetting is significantly reduced. Specifically, all versions of IT are no longer worse than their base versions, indicating that the ability to follow instructions is not as severely forgotten as in CPT. This is further supported by the MT-Bench scores, where we obtained 7.2031, 6.2094, and 7.3219 for IT-Mix, IT-In and IT-Gen, respectively, all of which are significantly better than the CPT counterparts.

We observe that IT-Mix is slightly better than other data versions, suggesting that mixing general tasks remains helpful to mitigating forgetting of general concepts and tasks, although the effect is much less pronounced compared to CPT. 
We also see that similar tasks improve significantly over base model while novel tasks (including financial tasks and general tasks) show little change. This indicates that, in contrast to CPT, domain has less impact in IT, but task generalization is a challenging issue. 

\begin{figure}[h]
\centering
\includegraphics[width=\columnwidth]{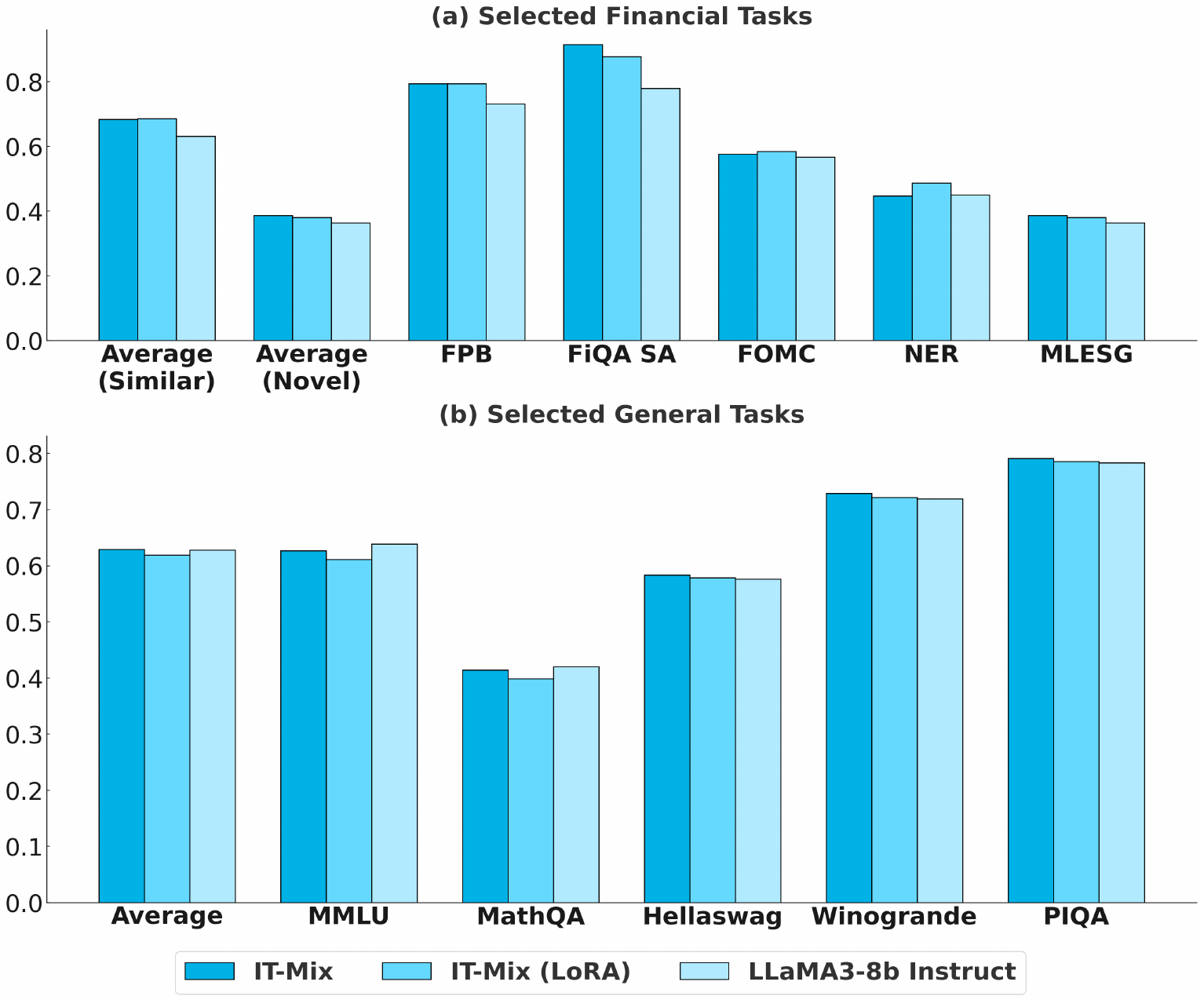}
 \vspace{-5mm}
\caption{Average performance on selected datasets for training Llama3-8b-instruct on IT-Mix {with full-model finetuning (IT-Mix) and LoRA finetuning (IT-Mix (LoRA))}. 
}
\label{fig.it_peft}
\vspace{-3mm}
\end{figure}

\paragraph{Comparison with LoRA.} 
\textcolor{black}{Another popular approach to adapt the LLM to specific domain is \textbf{Parameter-efficient Fine-tuning (PEFT)}, where the LLM parameters remain fixed, and only a small set of additional parameters are trained. This approach naturally mitigates forgetting issues and is more efficient in terms of trainable parameters. However, whether it can achieve performance comparable to full-model training is unclear. In Figure \ref{fig.it_peft}, we experiment with PEFT, specifically using LoRA~\citep{hu2021loralowrankadaptationlarge}, with a rank size of 128,\footnote{{Further decreasing or increasing the rank size did not show improvement in our preliminary experiments. For example, rank size of 32, 128 and 512 yield overall averages across 10 general tasks of 0.5267, 0.5331, and 0.5215, respectively, showing only minor differences.}}
 and compare its performance with full-model finetuning (IT-Mix). We observe that with and without LoRA performs similarly, confirming that LoRA is effective for task adaptation. However, the novel tasks still show little improvement, highlighting that task generalization still remains a significant challenge.}

A plausible reason for the lack of task generalization is that effective generalization may require exposure to a diverse range of tasks \citep{wei2022finetuned}, which is often impractical in certain domains, particularly long-tail ones. However, \textit{concepts} themselves may be inherently more generalizable due to the shared nature of concepts across tasks. Based on this, we propose adding CPT either before or concurrently with the IT stage and conduct training experiments accordingly.


\subsection{Combining CPT and IT}
\label{sec.joint_cpt_it}

A natural choice is to conduct CPT and IT sequentially \citep{lambert2024tulu3}. On the one hand, this is flexible as it allows for different settings (e.g., data size) in each stage. On the other hand, it does not help prevent forgetting during the CPT stage, leaving the LLM dependent on IT to `recover' its instruction-following capability. To make a more grounded decision, we conduct experiments on both sequential and joint training approaches. In joint training, an additional hyperparameter to consider is the mixture ratio. We \textit{down-sample} CPT data to match the size of IT data. In ``Other sampling strategies'' section, we will show this is the most effective strategy.

\begin{figure}[h]
\centering
\includegraphics[scale=11,width=\columnwidth]{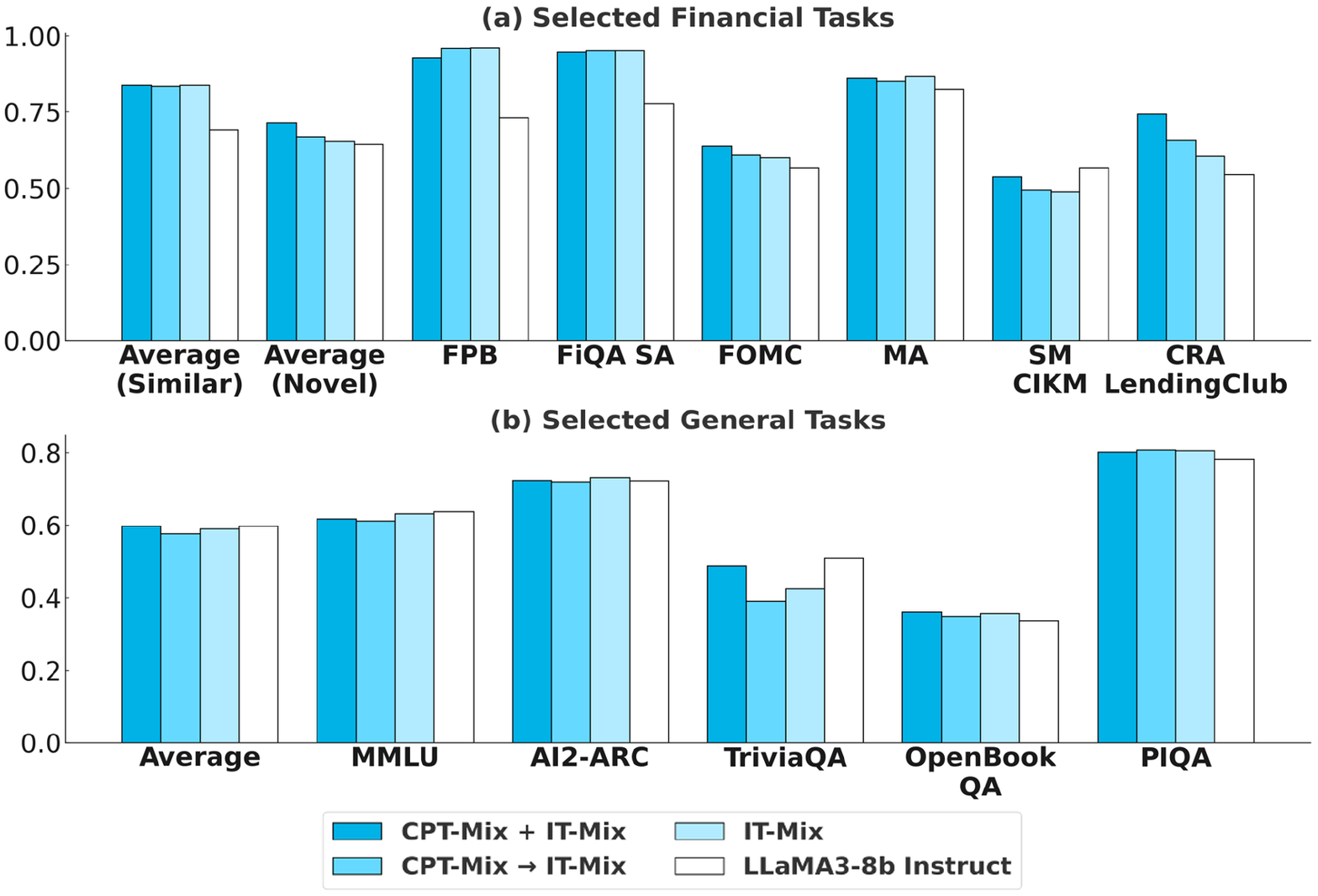}
\vspace{-8mm}
\caption{Average performance on selected datasets for training Llama3-8b-instruct on CPT-Mix and IT-Mix jointly (CPT-Mix + IT-Mix) and sequentially (CPT-Mix $\rightarrow$ IT-Mix). 
}
\label{fig.joint_sequential_combined}
\end{figure}

Figure~\ref{fig.joint_sequential_combined} illustrates the comparison between joint and sequential training. In both cases, different from IT-only results shown in Figure~\ref{fig.it_combined}, we see improved performance on similar and novel tasks. This supports our hypothesis that CPT can help improve the generalization of IT, as the concepts are likely able to be shared across different tasks. It is further interesting to see that even the general tasks are improved, indicating that there could be positive transfer between CPT and IT. Comparing the two, we observe that joint training outperforms sequential training across financial and general tasks, as well as similar and novel tasks, highlighting the importance of preventing forgetting of CPT and knowledge transfer between CPT and IT.

\begin{figure*}[t]
\centering
\includegraphics[width=\textwidth]{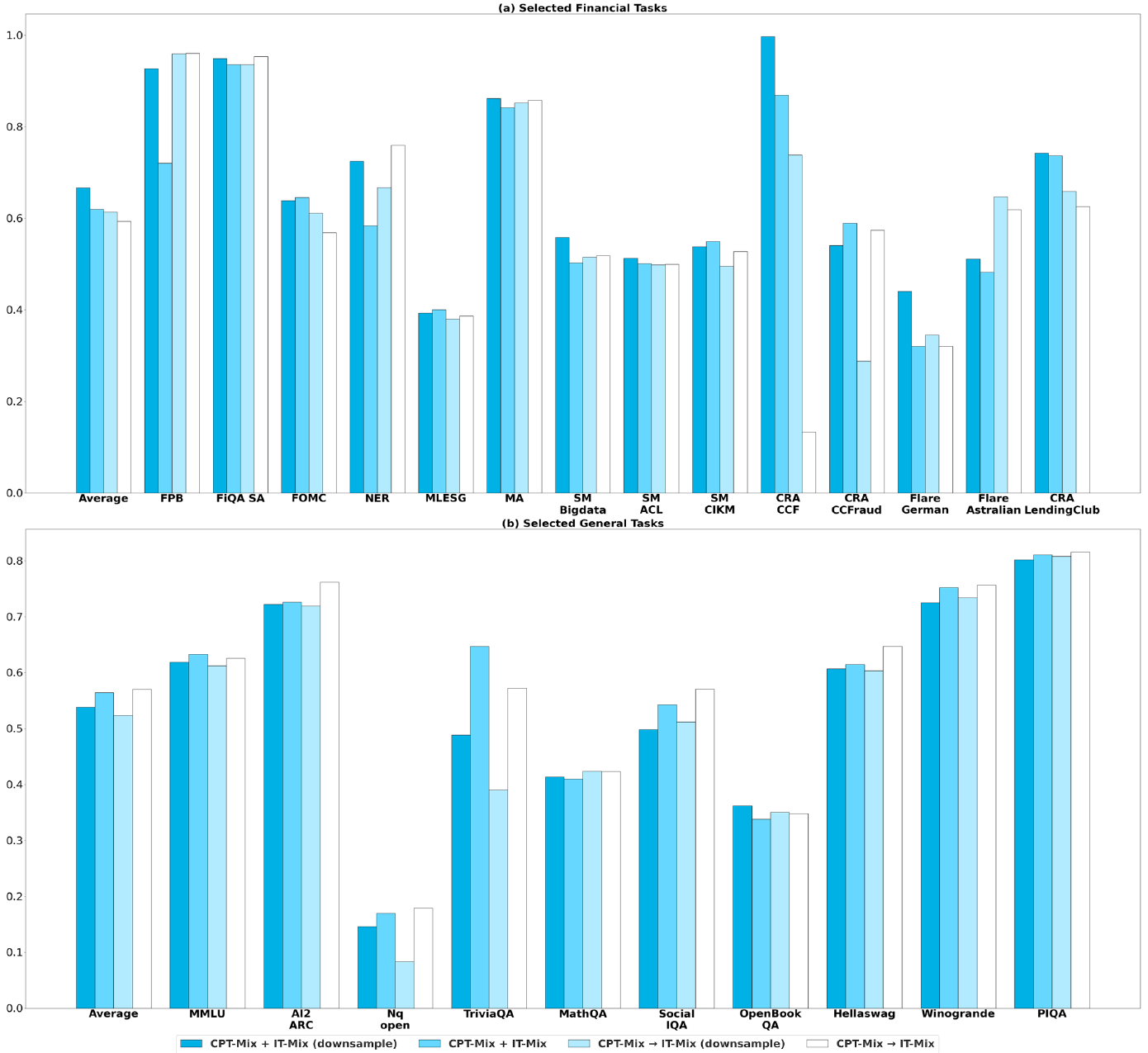}
\caption{Average and selected  datasets performance from down-sampling or no-sampling on CPT. 
}
\label{fig.downsample}
\vspace{-5mm}
\end{figure*}

\paragraph{Other Sampling strategies.}
Besides down-sampling, we also evaluate the performance under a `no-sampling' setting. Figure~\ref{fig.downsample} shows the results. We observe that in both joint and sequential training, down-sampling yields better results on financial tasks. This is understandable because down-sampling assigns more weight to IT, which is beneficial for the financial tasks. Interestingly, we observe the opposite trend for general tasks: no-sampling performs better. We hypothesize that this is because having more CPT data helps preserve general concepts more effectively, although it may diminish instruction-following abilities.


\begin{figure}[h]
\centering
\includegraphics[width=\columnwidth]{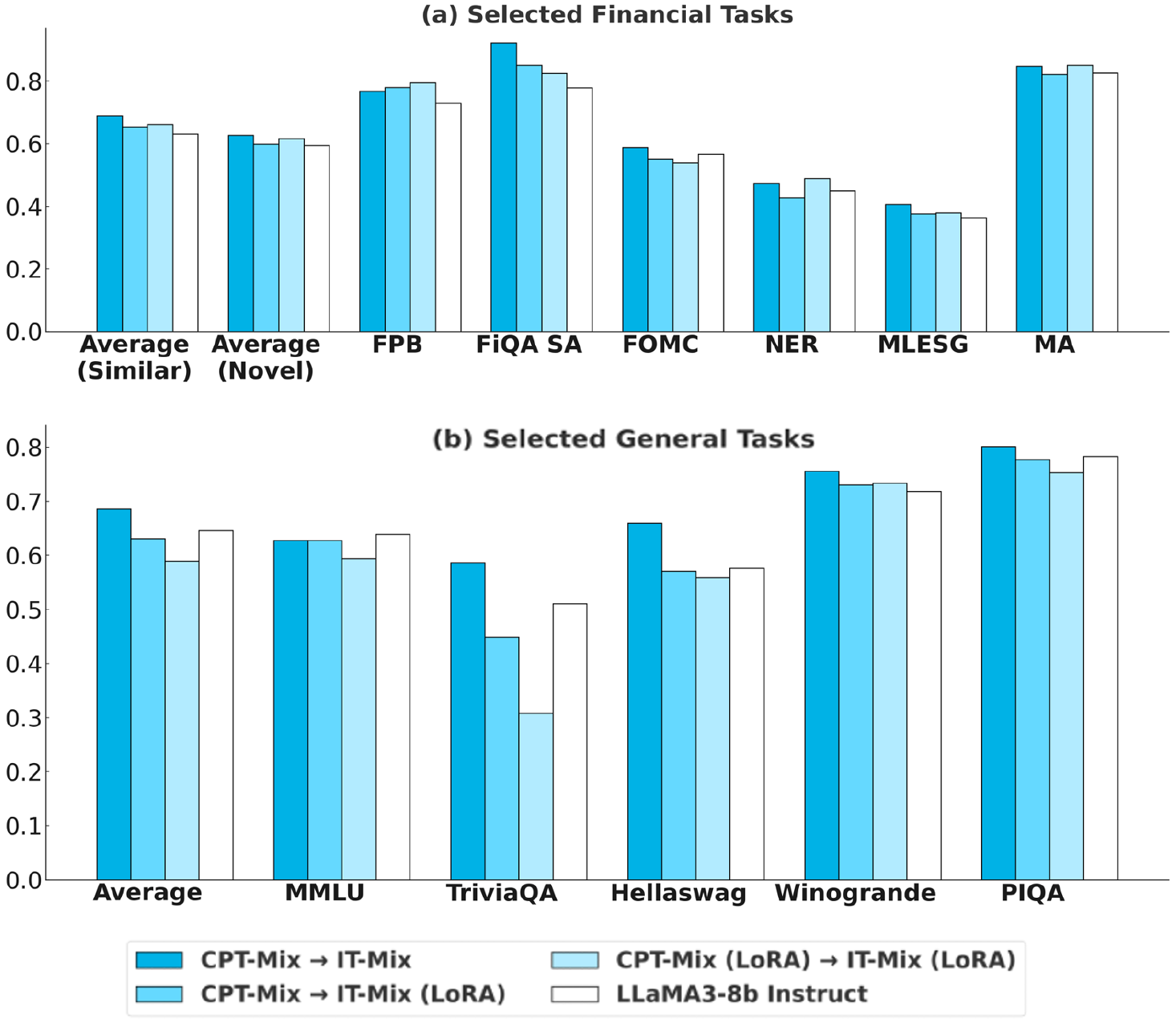}
\caption{Average performance on selected datasets for PEFT or full model fine-tuning for CPT and IT. 
}
\label{fig.joint_peft}
\end{figure}

\paragraph{Comparison with LoRA.}
\textcolor{black}{In Section~\ref{sec.it}, we showed that LoRA can effectively adapt tasks but still suffers from task generalization. While we already showed that CPT can help in full-model training setting, we now explore whether CPT can help in the PEFT setting as well. Figure~\ref{fig.joint_peft} presents the results of applying LoRA for IT and LoRA for both CPT and IT. Surprisingly, we find that full fine-tuning significantly outperforms the LoRA counterparts across similar and novel tasks.
This finding contrasts with our previous observations in Figure~\ref{fig.it_peft}, where performance with and without LoRA was comparable. Our results reveal that knowledge transfer from CPT to IT, which is crucial for task generalization, requires full-model training.}

\subsection{Preference Alignment}



\begin{figure}[t]
\centering
\vspace{-5mm}
\includegraphics[width=\columnwidth]{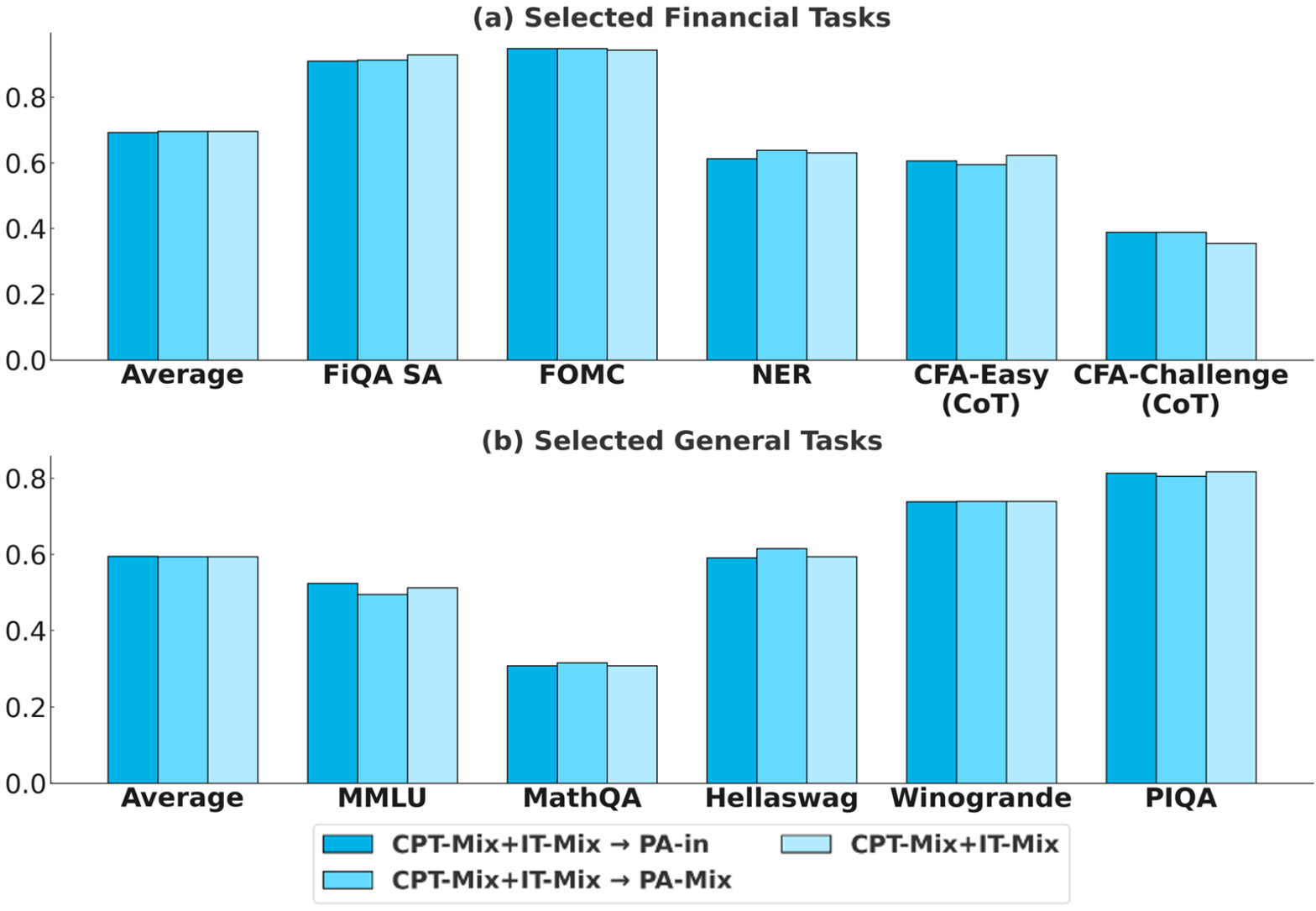}
\vspace{-8mm}
\caption{Average performance on selected datasets for PA training from the `CPT+IT' checkpoint on PA-Mix and PA-In.
}
\label{fig.pa_combined}
\vspace{-5mm}
\end{figure}

\textcolor{black}{\noindent\textbf{Negligible Forgetting in PA.}} As with CPT and IT, we begin by performing an ablation study on different data versions to evaluate their effectiveness.
Since the degree of forgetting diminishes from CPT to IT (as observed in \S\ref{sec.it}), we expect it to be even less pronounced in PA.  
To quickly evaluate this hypothesis, we {take a naive approach and create \textit{PA-Mix} and \textit{PA-In} } by using either the provided or GPT4o generated responses (as done for IT in \S\ref{sec.it}) as the `chosen' samples and the output of `CPT+IT' checkpoint as the `rejected' ones{, based on the prompts of IT-Mix and IT-In, respectively.} 

Figure~\ref{fig.pa_combined} shows the results {after PA training} for PA-In and PA-Mix from the `CPT+IT' checkpoint.\footnote{PA trained from Llama3-8b-instruction has shown worse results compared to training from the `CPT+IT' checkpoint in our preliminary experiments, as PA requires a strong initialization checkpoint. For instance, PA-Mix from Llama3-8b-instruction achieves only 29.99 on EDTSUM, whereas the CPT+IT' counterpart achieves 54.21. As a results, we only investigate training PA from `CPT+IT' checkpoint. } We observe that PA-In performs comparably to PA-Mix, indicating that it may not be essential to include general tasks to prevent forgetting of concepts or tasks, unlike the cases of CPT and IT. {This suggests that PA training can focus on in-domain tasks, without requiring a broader set of general tasks or raising concerns about forgetting. Given this, we use CFA exams (Table~\ref{tab.prompt_curation} in \S\ref{sec.ablation_fin_train}) as a representative source for in-domain reasoning because they cover diverse financial scenarios, emphasize complex reasoning, and, most importantly, are derived from real-world exams. These characteristics make them a strong proxy for a broader range of financial tasks, ensuring that the model generalizes effectively within the financial domain while simplifying the training process.
}


Another crucial observation is that there is not much difference even for \textcolor{black}{unseen similar} 
tasks (FiQA SA, FOMC and NER) and reasoning tasks (CFA-Easy and CFA-Challenge). This highlights the limitations of the current naive PA approach and suggests room for further improvement. \textcolor{black}{In \ourframework, we propose a novel PA approach that constructs preference data guided by both outcome and process reward signals.}

\section{Other Capabilities}
\label{sec.extra_capability}

Besides those core capabilities mentioned in \S\ref{sec.perspectives}, domains may vary significantly in their \textit{sensitivity}. For instance, the medical domain is highly sensitive, requiring utmost accuracy and strict adherence to ethical considerations. In contrast, domains such as entertainment may have more relaxed requirements. Another important consideration is \textit{multi-modality}, as some domains require handling multiple types of input and output formats. For example, the healthcare domain may involve processing medical images alongside textual reports, while the e-commerce domain may integrate product descriptions, images, and customer reviews into a unified response. Similarly, scientific research often combines charts, graphs, and textual analysis to present findings effectively.

\section{FinTrain}
\label{sec.ablation_fin_train}

\begin{table}[h]
\setlength{\tabcolsep}{1pt}
\centering
\resizebox{\columnwidth}{!}{
\begin{tabular}{lllll}
\toprule
\textbf{Capability} & \textbf{Domain} & \textbf{  Dataset} & \textbf{Size} & \textbf{Reference} \\
\toprule 
 Concept & General & NaturalInstrution & 100,000 &  \citet{naturalinstructions} \\
&  & PromptSource & 100,000 & \citet{bach2022promptsource} \\
&  & Math & 29,837 & \citet{amini-etal-2019-mathqa} \\
&  & Aqua & 97,500 & \citet{ling2017program} \\
&  & CREAK & 10,200  & \citet{onoe2021creak} \\
&  & ESNLI & 549,367  & \citet{NIPS2018_8163} \\
&  & QASC & 8,130  & \citet{allenai:qasc} \\
&  & SODA & 1,190,000  & \citet{kim2022soda} \\
&  & StrategyQA & 2,290 & \citet{geva2021strategyqa} \\
&  & UnifiedSKG & 779,000 & \citet{UnifiedSKG} \\
&  & GSM8K & 7,470 & \citet{cobbe2021gsm8k} \\
&  & ApexInstr & 1,470,000  & \citet{Huang2024OpenCoderTO} \\
&  & DeepmindMath & 379,000 & \citet{2019arXiv} \\
&  & DialogueStudio & 1,070,000  & \citet{zhang2023dialogstudio} \\
 \rowcolor[HTML]{D7F6FF} 
 & Finance & \textbf{{\color[HTML]{00B2E6} Fineweb-Fin}} & 4,380,000 & - \\
\rowcolor[HTML]{D7F6FF} 
&  & \textbf{{\color[HTML]{00B2E6} Book-Fin}} & 4,500 & - \\
\hline
\textit{Total} & & & 10,177,294 & 
\\
\bottomrule
\end{tabular}
}
\caption{Summary of curated texts. New datasets released with \ourframework 
are color-highlighted for emphasis. 
}
\label{tab.text_curation}
\end{table}

\begin{table*}[h]
\centering
\resizebox{0.6\textwidth}{!}{
\begin{tabular}{llllll}
\toprule
\textbf{Capability} & \textbf{Domain}  & \textbf{Task} & \textbf{ Dataset} & \textbf{Size} & \textbf{Reference} \\
\toprule 
 Tasks & Finance & Relation Cls. & FingptFinred & 27,600 & \citet{sharma2022finred} \\
 & & NER & FingptNERCls & 13,500 & \citet{yang2023fingpt} \\
 &  &   & FingptNER & 511 & \citet{salinas-alvarado-etal-2015-domain} \\
 & & Headline Cls. & FingptHeadline & 82,200 & \citet{DBLP:journals/corr/abs-2009-04202} \\
 & & Sentiment Cls. &  SentimentCls  &  47,600 & \citet{yang2023fingpt} \\
 & & & SentimentTra & 76,800 & \citet{yang2023fingpt} \\
 & & Summariz. & TradeTheEvent & 258,000 & \citet{zhou-etal-2021-trade} \\
IF/Chat & General & IF/Chat & SelfInstruct & 82,000 & \citet{selfinstruct}  \\
 & & & SlimOrca  & 518,000 & \citet{SlimOrca}  \\
 & & & UltraChat & 774,000 & \citet{ding2023enhancing} \\
 & & & ShareGPT & 100,000 & \href{https://huggingface.co/datasets/anon8231489123/ShareGPT_Vicuna_unfiltered}{Link}
 \\
  & Finance & QA & FinanceInstruct & 178,000 & \href{https://huggingface.co/datasets/sujet-ai/Sujet-Finance-Instruct-177k}{Link} \\
 &  && FingptConvfinqa & 8,890 & \citet{chen-etal-2022-convfinqa} \\
& & & FlareFinqa & 6,250 & \citet{chen-etal-2021-finqa} \\
& & & FlareFiqa & 17,100 & \citet{yang2023fingpt} \\
Reasoning & Math & QA & OrcaMath & 200,000 &  \citet{mitra2024orcamath}
\\
 & &  & MetaMathQA & 395000 &  \citet{yu2023metamath}\\
  & &  & MathInstruct  & 262,000 & \citet{yue2023mammoth} \\
& Code & QA & MagicodeInstruct & 111,000 & \citet{luo2023wizardcoder} \\
\rowcolor[HTML]{D7F6FF} 
  & Finance &  CFA Exam & \textbf{{\color[HTML]{00B2E6} Exercise}} & 2,950 & - \\
\textit{Total} & & & & 3,161,401 & 
\\
\bottomrule
\end{tabular}
}
\caption{Summary of our curated prompts. New datasets released with \ourframework 
are color-highlighted for emphasis. For datasets without formal references but only a URL, we provide their links.
}
\label{tab.prompt_curation}
\end{table*}

\paragraph{Continual Pre-training Text Curation}
\label{sec.text_curate}
To introduce domain concepts while preserving general concepts, we curate texts for CPT. Table~\ref{tab.text_curation} summarizes the texts curation datasets. 
Specially, for general concepts, research has shown that a `small' amount of general text (as little as 1\%) can effectively mitigate the forgetting issue \citep{scialom-etal-2022-fine}. Therefore, we focus on collecting a relatively small but high-quality set of general-domain text. To achieve this, we use \textit{verifiable text}, which is text written by humans and previously used in supervised tasks in the literature. Note that this contrasts with using \textit{unverifiable} web text such as C4 \citep{2020t5}.


For domain concept, our goal is to collect both a large volume of data and maintain high quality. Following practices from the literature on training general LLMs \citep{lambert2024tulu3,gunasekar2023textbooksneed}, 
we source financial texts from primarily relevant websites and books. 
Specifically, we source financial text from two primary resources. The first source is \textit{web text}, where we filter non-financial content from the FineWeb using URLs like `\url{sec.gov}' and `\url{investopedia.com}'. The second source is \textit{books}. We manually select 10 finance-related topics (e.g., `economics' and `management'), download books on these topics, and convert them to text using OCR~\citep{textract}. Since OCR can make mistakes, we further employ a strong LLM to filter out content lacking educational value or unrelated to finance. Details on the financial URLs, finance-related topics, and the prompts used for filtering is shown below:


\textbf{$\bullet$ Selected Financial URLs.} We curated a selection of \textit{70} financial websites to comprehensively cover diverse aspects of finance-related content on the web. These include trusted sources from financial institutions, regulatory agencies, educational platforms, and industry-specific news outlets. This diverse collection ensures representation across sub-domains such as investment, banking, personal finance, regulatory compliance, and financial planning, offering a well-rounded foundation that can cover most of the finance content in the web.


\textbf{$\bullet$ Selected Topics.} We select \textit{12} topics that are cover most of books in finance. \textit{5} of them are from \textit{business} areas, including \textit{business}, \textit{Accounting}, \textit{Accounting}, \textit{Management}, \textit{Marketing}, \textit{Trading}; \textit{1} is from \textit{Mathematics}, i.e., \textit{Mathematical Economics}; and \textit{4} are from \textit{Economy} area, including \textit{economy}, \textit{econometrics}, \textit{investing}, and \textit{markets}. We crawled the web to downloaded the books from the corresponding topics. For CFA, we use the material provided by CFA prep providers.

\textbf{$\bullet$ Prompt for Filtering the Text.} We explored various prompt formats to automatically extract an financial score using an LLM and found that the additive scale by \citet{yuan2024selfrewarding} worked best. Figure~\ref{fig.prompt_filter_text} shows the prompt we used to filter the `low-quality' text. Specifically, this prompt allows the LLM to reason about each additional point awarded, unlike the single-rating Likert scale which fits samples into predefined boxes. Then, to avoid the LLM favoring highly technical content like academia papers, we focused on financial student level knowledge. By setting a threshold of 4 (on a scale of 0 to 5) during the filtering process, we were able to also retain some high-quality financial content. 

\begin{figure*}[h]
\centering
\begin{tcolorbox}[colback=outerboxcolor,colframe=innerboxcolor,title=Prompt for Filtering the Text,fonttitle=\bfseries,arc=3mm,boxrule=1pt]
Below is an extract from a text book. Evaluate whether the book has a high financial value and could be useful in an financial setting for teaching financial students using the additive scoring system described below. Points are accumulated strictly based on the satisfaction of each criterion:

\begin{itemize}
    \item Add 1 point if the extract provides educational value for financial students whose goal is to learn financial concepts or take financial exams. It is acceptable if quizzes are not included; however, if quizzes are present, detailed solutions and explanations must also be provided.
    \item Add another point if the extract addresses certain elements pertinent to finance and aligns closely with financial standards. It might offer a superficial overview of potentially useful topics or present information in a disorganized manner and incoherent writing style.
    \item Award a third point if the extract is appropriate for financial use and introduces key concepts relevant to financial curricula. It is coherent and comprehensive.
    \item Grant a fourth point if the extract is highly relevant and beneficial for financial learning purposes for a level not higher than financial students, exhibiting a clear and consistent writing style. It offers substantial financial content, including exercises and solutions, with minimal irrelevant information, and the concepts aren't too advanced for financial students. The content is coherent, focused, and valuable for structured learning.
    \item Bestow a fifth point if the extract is outstanding in its financial value, perfectly suited for teaching either at financial students. It follows detailed reasoning, the writing style is easy to follow and offers profound and thorough insights into the subject matter, devoid of any non-financial or complex content.
\end{itemize}
The extract: \texttt{<EXAMPLE>}.
\\\\
After examining the extract, You will output a json object containing the following 2 fields:
\begin{lstlisting}[style=jsonstyle]
{
    "Justification": string // Briefly justify your total score, up to 100 words.
    
    "Score": integer // Conclude with the score 
}
\end{lstlisting}

\end{tcolorbox}
\caption{Prompt for filtering the text}
\label{fig.prompt_filter_text}
\end{figure*}


\paragraph{Insturction Prompt Curation}
\label{sec.prompt_curate}
Prompts represent the diverse ways users may interact with models and serves the essential component for IT and PA. Table~\ref{tab.prompt_curation} summarizes the prompts curation datasets. Specifically, we conduct a broad survey and source general, financial, instruction-following, and reasoning tasks from \textit{public datasets}. 
To promote \textit{diversity}, we include datasets like Flare-FinQA ~\citep{chen-etal-2021-finqa}, a large open QA dataset in finance, and UltraChat~\citep{ding2023enhancing}, a dataset shown to perform well for IT in the literature \citep{tunstall2024zephyr,ivison2024unpackingdpoppodisentangling}. 
Additionally, we find that exercises or demonstrations from books that were curated in \S\ref{sec.text_curate} is valuable for reasoning tasks as they usually involve challenging reasonings and come with ground truth answers and sometimes even include human-written chain-of-thought (CoT) explanations. 

Figure~\ref{fig.prompt_extract_exercise} shows the prompt we used to extract exercises from books. We carefully design the prompt to extract both the question part of an exercise, which potentially include questions, scenario and exhibits, and the answer part of the exercise, which may include answer choices and solution. In books, questions and their corresponding answers can be located far apart (e.g., the questions may appear at the beginning while the solutions are provided at the end), meaning they may not be captured within the same chunk. As a result, some questions may not have corresponding extracted answers. For such cases, GPT-4o's generated answers are used when converting the prompt into instruction-following or preference-alignment data.

\begin{figure*}[ht]
\centering
\begin{tcolorbox}[colback=outerboxcolor,colframe=innerboxcolor,title=Prompt for Extracting Exercise from Book,fonttitle=\bfseries,arc=3mm,boxrule=1pt]
You are an educational assistant aims to extract all questions from the provided material. Look for specific indicators such as "example," "quiz," "questions," or similar terms to identify where the questions are located. If the material includes scenarios or exhibits, must include all details related to them. Do not create or derive any questions or come up with content on your own—strictly extract what is present in the material. Make sure no question is missed. If one scenario or exhibits corresponds to multiple questions, duplicate the scenarios and exhitbits so that the number of questions match the number of scenarios and exhibits. \\

The material: \texttt{<MATERIAL>}.\\

After performing these tasks, You will output a json object containing the following fields:
\begin{lstlisting}[style=jsonstyle]
{
  "Justification": "string", // A brief justification for your extractions,  up to 100 words. 
  
  "Questions": "string", // A list of questions extracted from the material. Only extract the exact questions presented in the text. 
  
  "Scenario": "string", // A list of scenarios corresponding to the above questions.  If the material does not provide the scenario place "N/A."  Do not do any derivation or reference, must output the exact same, detailed and complete scenarios. The scenario may contain multiple paragraphs or even splited by the exhibits, combine them into one string. The scenario can be long, you may modify it to make it shorter,  but must not change its meaning.  
 
  "Exhibit": "string", // A list of exhibits or tables corresponding to the above questions. If the material does not provide the exhibit, place "N/A." Do not do any summary, or derivation or cutting,  must output the exact same, detailed and complete exibits. There may be multiple exhibits involved in a scenario, combine them into one string. The exhibit can be long, you may modify it to make it shorter. Must keep the table format 

  "Answer Choices": "string", // A list of answer choices corresponding to the above questions. If the material does not provide answer choices, place "N/A." 
  
  "Answer": "string" // A list of answers corresponding to the above questions. Answers should only be included if provided in the material. If no answer is given, place "N/A." If explanations or reasoning steps or equations are included, must capture all of them. Must not answer it yourself if there is no answer provided in the material. Make sure the final number of questions equals to number of scenario equals to number of exhibits equals to number of answers
}
\end{lstlisting}

\end{tcolorbox}
\caption{Prompt for extracting exercises from books}
\label{fig.prompt_extract_exercise}
\end{figure*}

\section{FinEval}
\label{sec.ablation_fin_eval}

\begin{table*}[t!]
\centering
\setlength{\tabcolsep}{2pt}
\resizebox{0.7\textwidth}{!}{
\begin{tabular}{lllllllll}
\toprule
\textbf{Capability} & \textbf{Domain}  & \textbf{Task} 
 & \textbf{Evaluation Dataset} & \textbf{Size} & \textbf{Reference} \\
\toprule 
\multicolumn{6}{c}{\textbf{Unseen - Similar}} &  \\
Tasks & Finance & Sentiment Analysis & FPB & 970 &  \citet{Malo2014GoodDO}\\
&  & & FiQA   SA & 235 &  \citet{maia201818}\\
 && Monetary policy Stance & FOMC & 496 & \citet{shah-etal-2023-trillion} \\
 && Named entity recognition & NER & 98 &  \citet{salinas-alvarado-etal-2015-domain}\\
 && Abstractive Summarization & EDTSUM & 2,000 & \citet{zhou-etal-2021-trade} \\
\textit{Total} & &  & & 3,799 &  \\
\multicolumn{6}{c}{\textbf{Unseen - Novel}} &  \\
Concept & General & Knowledge Recall& MMLU & 14,042 & \citep{hendryckstest2021} \\
 &  & & AI2-ARC & 3,548 & \citet{Clark2018ThinkYH}  \\
&  &  & Nq-open & 7,842 & \citet{nq-open} \\
\rowcolor[HTML]{D7F6FF} 
& Finance & & \textbf{{\color[HTML]{00B2E6}MMLU-Finance}} & 1,460 & - \\
Tasks & Finance & Extractive Summarization & Flare-ECTSUM & 495 & \citet{mukherjee-etal-2022-ectsum} \\
&  & ESG Issue Classification & MLESG & 300 & \citet{chen-etal-2023-multi-lingual} \\
&  & Rumour Detection & MA & 500 & \citet{Yang2020GeneratingPC} \\
&  & Stock Movement Prediction & SM-Bigdata & 1,470 & \citet{soun2022accurate} \\
&  & & SM-ACL & 3,720 & \citet{xu-cohen-2018-stock} \\
&  & & SM-CIKM & 1,140 & \citet{WuCIKM18} \\
&  & Fraud Detection & CRA-CCF & 2,280 & \citet{feng2024empoweringmanybiasingfew}\\
& & & CRA-CCFraud & 2,100 & \citet{feng2024empoweringmanybiasingfew} \\
&  & Credit Scoring & Flare-German & 200 &  \citet{statlog_(german_credit_data)_144}\\
&  & & Flare-Astralian & 139 & \citet{statlog_(australian_credit_approval)_143} \\
&   & & CRA-LendingClub & 2,690 & \citet{feng2024empoweringmanybiasingfew} \\
&  & Distress Identification & CRA-Polish & 1,740 & \citet{feng2024empoweringmanybiasingfew} \\
&  &  & CRA-Taiwan & 1,370 & \citet{feng2024empoweringmanybiasingfew} \\
& & Claim Analysis & CRA-ProroSeguro & 2,380 & \citet{feng2024empoweringmanybiasingfew} \\
&  & & CRA-TravelInsurance & 2,530 &  \citet{feng2024empoweringmanybiasingfew}\\
&  & Tabular QA & Flare-TATQA & 1,670 & \citet{zhu-etal-2021-tat} \\
&  & Open QA & Finance Bench & 150 & \citet{islam2023financebench} \\
IF/Chat & General & Precise IF & MT-bench & 80 & \citet{zheng2023judging} \\
 Reasoning & Math & Reasoning & MathQA & 2,985 & \citet{amini2019mathqa} \\
& General & Social Reasoning & Social-IQA & 2,636 & \citet{Welbl2017CrowdsourcingMC} \\
 &  &Common Reasoning & Open-book-qa & 500 & \citet{OpenBookQA2018} \\
 &  && Hellaswag & 10,003 &  \citet{zellers2019hellaswag}\\
&  && Winogrande & 1,767 &  \citet{sakaguchi2019winogrande}\\
&  & & PIQA & 3,000 & \citet{Bisk2020} \\
& Finance &Exam & CFA-Easy & 1,030 & \href{https://huggingface.co/datasets/ChanceFocus/flare-cfa}{Link} \\
\rowcolor[HTML]{D7F6FF} 
& & & \textbf{{\color[HTML]{00B2E6}CFA-Challenge}} & 90 & - \\
\textit{Total} & & & & 91,872 & 
\\
\bottomrule
\end{tabular}
}
\caption{Summary of our evaluation dataset. New datasets released with \ourframework are color-highlighted for emphasis. 
}
\vspace{-3mm}
\label{tab.eval_data}
\end{table*}

With the breakdown of capabilities in \S\ref{sec.perspectives}, our evaluation framework consists of a suite for assessing these capabilities using development sets and unseen (held-out) evaluation sets. Our development set is directly split from the training data at each stage. Table~\ref{tab.eval_data} outlines the capabilities and the evaluation benchmarks selected to cover these capabilities. {Crucially, we did not examine scores on our unseen set while developing the models, which allows us to observe how much we may have overfitted to particular evaluations in our decisions around training recipe.} For the unseen tasks (Table~\ref{tab.eval_data}), we manually review each individual dataset and have the following considerations.


\noindent\textbf{$\bullet$ Benchmarking tasks.} Corresponding to the capabilities, we consider a diverse set of benchmarking tasks. For \textit{concepts}, we include knowledge tasks in the general domain, such as AI2-ARC \citep{Clark2018ThinkYH}, as well as in  finance, such as MMLU-Finance \citep{hendryckstest2021}. For \textit{tasks}, we consider general tasks, such as Social-IQA \citep{Welbl2017CrowdsourcingMC}, and domain-specific tasks, such as MLESG \citep{chen-etal-2023-multi-lingual}. Notably, we intentionally include a few financial tasks such as Flare-TATQA \citep{zhu-etal-2021-tat} and SM-Bigdata \citep{soun2022accurate} that require understanding of tabular data, as this data format is common in this domain. For \textit{IF/Chat capabilities}, we utilize popular instruction-following benchmarks, such as MT-Bench \citep{zheng2023judging}. For \textit{reasoning}, we include general reasoning tasks, such as MathQA \citep{amini2019mathqa} and Hellaswag common sense reasoning \citep{zellers2019hellaswag}, as well as domain-specific reasoning tasks, such as CRA-ProroSeguro claim analysis \citep{feng2024empoweringmanybiasingfew}. We also construct a new benchmark on CFA-Challenge based on CFA Level III, one of the most challenging financial exams that requires comprehensive reasoning \citep{labs2024thalletexthyperlocallyaugmented,callanan-etal-2024-gpt}.

\noindent\textbf{$\bullet$ Evaluation method.} 
We split our evaluation set into two types based on their exposure to \textit{Instruction tuning (IT) data} (Table~\ref{tab.eval_data}). The first type, \textit{Similar}, includes tasks \textcolor{black}{whose types have been encountered during training, even if the specific tasks themselves are unseen (e.g., a new NER task). The second type, \textit{Novel}, includes tasks whose types have not been seen during training, representing entirely new challenges for the model (e.g., stock movement prediction).} 
We use two different evaluation methods based on the nature of the benchmarks. For knowledge and NLP tasks (e.g., NER), we employ a straightforward \textit{direct answer} evaluation. For reasoning tasks (e.g., CFA-Challenge), we use a \textit{0-shot chain-of-thought (CoT) \citep{wei2023chainofthoughtpromptingelicitsreasoning} answer} evaluation to enhance the reliability of our evaluation. This also exposes the reasoning path, allowing us to investigate the causes of incorrect answers and enabling a more fine-grained comparison across different models.

\section{Preliminary Experiments on Older Financial LLMs}
\label{ap.prelim_baseline}

As mentioned in Section~\ref{sec.experiment}, the reason we did not include older financial LLMs is that they are either not publicly available (e.g., Bloomberg GPT) or clearly worse than our model. As a result, we only include the SoTA finance LLM (i.e., Palmyra Fin 70b) in the comparison.

To further support this point, we compare performance on overlapping evaluation benchmarks, using the reported numbers for other baselines extracted from their papers. 
We made careful efforts to ensure comparability:

\textbf{$\bullet$  Metrics.} We noticed that different metrics were used across baselines and our methods. For example, some baselines reported F1 scores for FPB and FiQA SA, while we originally reported accuracy. For NER, the baselines used Entity F1, whereas we initially reported ROUGE scores. To ensure fair comparison, we re-ran our evaluation using the same metrics. We reported both accuracy and F1 for FPB and FiQA SA, and used Entity F1 for NER.

\textbf{$\bullet$ Test Datasets.} The test datasets are the same. We follow the datasets used in \citet{xie2023pixiulargelanguagemodel}\footnote{\url{https://huggingface.co/collections/TheFinAI/english-evaluation-dataset-658f515911f68f12ea193194}}, which include 235 test samples for FiQA SA, 970 samples for FPB, and 98 samples for NER. These statistics also match those reported in Table E.1 of our Appendix. We do not use the training or validation sets, as our evaluation is conducted purely in the zero-shot setting. The baseline results are taken directly from Table 5 in \citet{xie2023pixiulargelanguagemodel}, which ensures consistency in comparison and also corresponds to Table 1 in \citet{xie-etal-2024-efficient}.

Table~\ref{tab.older_baseline} shows the results. These results clearly show that our model outperforms these older financial LLMs, including significantly larger models such as FinMA 30B. Moreover, their reported results are based on few-shot settings, whereas our evaluations are conducted in the zero-shot setting, further highlighting the effectiveness of our approach.

\begin{table}[t]
\centering
\setlength{\tabcolsep}{2pt}
\resizebox{\columnwidth}{!}{
\begin{tabular}{ll>{\columncolor{lightblue}}lllll}
\toprule
Dataset & Metric & \begin{tabular}[c]{@{}l@{}}Llama-Fin \\ 8b\end{tabular} & \begin{tabular}[c]{@{}l@{}}Bloomberg\\GPT\end{tabular} & \begin{tabular}[c]{@{}l@{}}FinPythia \\ 7B\end{tabular} & \begin{tabular}[c]{@{}l@{}}FinMA \\ 7B\end{tabular} & \begin{tabular}[c]{@{}l@{}}FinMA \\ 30B\end{tabular} \\
FPB & Acc & \textbf{91.13} & --- & 59.90 & 86.00 & 87.00 \\
 & F1 & \textbf{91.28} & 51.07 & 64.43 & 86.00 & 88.00 \\
FiQA SA & Acc & \textbf{95.32} & --- & 52.34 & 84.00 & 87.00 \\
 & F1 & \textbf{95.39} & 75.07 & 53.04 & --- & --- \\
NER & EntityF1 & \textbf{77.09} & 60.82 & 48.42 & 75.00 & 62.00\\
\bottomrule
\end{tabular}
}
 \vspace{-3mm}
\caption{Experiments on older baselines.
}
\vspace{-6mm}
\label{tab.older_baseline}
\end{table}

\section{Summary of the Final Recipe and Hyper-parameters}
\label{sec.hyper_parameter}

\begin{table*}[h]
\centering
\begin{tcolorbox}[colback=outerboxcolor,colframe=innerboxcolor,title=Final Recipe for \ourmodel,fonttitle=\bfseries,arc=2mm,boxrule=1pt]
\resizebox{\textwidth}{!}{
\begin{tabular}{lll}
\toprule
\multicolumn{3}{l}{\textbf{Continual Pre-training (CPT) and Instruction Tuning (IT)}} \\
\toprule
\textbf{Data} & 50\% CPT, 50\% IT &  \\
\textbf{Curriculum} & \textbf{Group 1} & \textbf{CPT:} 50\% Domain-specific Text (Web and book), 50\% General text (verfiable text) \\
 &  & \textbf{IT:} 20\% Domain-specific tasks, 80\% General tasks \\
 & \textbf{Group 2} & \textbf{CPT:} Group 1 data + domain-specific books \\
 &  & \textbf{IT:} Group1 + Exercises extracted from books \\
\textbf{Steps} &  & \begin{tabular}[c]{@{}l@{}}\textbf{Group 1:} 3.84B tokens; \textbf{Group 2:} 1.66B tokens \\ (8,000 context length, 16 A100)\end{tabular} \\
\textbf{Model} & \textbf{Intialization} & Llama3-8b-instruct \\
 & \textbf{Attention} & \textbf{CPT:} full attention with cross-document attention masking \\
 &  & \textbf{IT:} attention with instruction mask-out and cross-document attention masking \\
\textbf{Optim.} &  & AdamW (weight decay = 0.1, $\beta_1$=0.9, $\beta_2$=0.95) \\
 & \textbf{LR} & \textbf{Group 1:} 5e-6 with 10\% warmup; \textbf{Group 2:} 5e-6 with 50\% warmup \\
 & \textbf{Batch size} & 128K tokens \\
  \textbf{Stop Cri.} &  Loss of development set stops decreasing ($\approx$ 1 epoch) \\
 \toprule
\multicolumn{2}{l}{\textbf{Preference Alignment (PA)}} &  \\
\toprule
\textbf{Data} & \multicolumn{2}{l}{FAP and SCP } \\
\textbf{Steps} & 24.58 M tokens &  \\
\textbf{Model} & \textbf{Initialization} & CPT+IT \\
 & \textbf{Loss} & DPO with an additional
negative log-likelihood term\\
& \textbf{Attention} & Attention with instruction mask-out and cross-document attention masking \\

\textbf{Optim.} & \textbf{LR} & 5e-7 with 10\% warmup\\
 & \textbf{Batch size} & 32K tokens \\
 \textbf{Stop Cri.} & Loss of development set stops decreasing
\end{tabular}
}
\end{tcolorbox}
\caption{Final recipe of \ourmodel. The joint training of CPT and IT is structured into two groups, with each group undergoing joint training sequentially. The second group utilizes higher-quality data (sourced from books), following the typical curriculum training practice~\citep{gao2024trainlongcontextlanguagemodels}. For PA, we employ a modified DPO loss with an additional
negative log-likelihood term, similar to \citet{pang2024iterative}, as it has shown to be more effective than relying solely on the original DPO loss. }
\label{tab.final_recipe}

\end{table*}



\section{GenRM Prompt Details}
\label{sec.gen_rm_detail}

In Figure~\ref{fig.reasoning}, we simplified the prompt for GenRM for the purpose of illustration. In this section, we give full detailed of the prompt for Final Answer Preference (FAP) and Stepwise Corrective Preference (SCP) in Figure~\ref{fig.prompt_for_genorm} and Figure~\ref{fig.prompt_for_genprm}, respectively.


\begin{figure*}[ht]
\centering
\begin{tcolorbox}[colback=outerboxcolor,colframe=innerboxcolor,title=Prompt for FAP,fonttitle=\bfseries,arc=3mm,boxrule=1pt]

You given a question, a reference answer and a proposed answer, you task is to determine the correctness of the proposed answer. First, extract the final answer (for example, A, B or C) from the reference answer. Second, extract the final answer from the proposed answer (for example, A, B or C). Finally, compare the two final answer to determine the correctness. Do not do any extra reasoning, must determine the correctness soley based on the given reference and proposed answer.\\

Question: \texttt{<QUESTION>}\\
Reference Answer: \texttt{<REFERENCE>}\\
Proposed Answer: \texttt{<PROPOSAL>}\\

After performing these tasks, You will output a json object containing the following fields:

\begin{lstlisting}[style=jsonstyle]
{
  "Justification": "string", // A brief justification for your output,
  up to 100 words.

  "Correctness": "string", // If the proposed answer has the same final final answer as the reference answer (for example, both choose A or have the same answer),  output 'correct'.  Put 'wrong' to all other cases. For example, if the proposed answer has a different final answer comparing to the reference answer, put 'wrong'. If the proposed answer does not explicitly give a final answer to the question, put 'wrong'. If the proposed answer gives more than one final answer to the question, put 'wrong'.
}
\end{lstlisting}

\end{tcolorbox}
\caption{Prompt for FAP}
\label{fig.prompt_for_genorm}
\end{figure*}

\begin{figure*}[ht]
\centering
\begin{tcolorbox}[colback=outerboxcolor,colframe=innerboxcolor,title=Prompt for Prompt for SCP,fonttitle=\bfseries,arc=3mm,boxrule=1pt]

Given a question, a reference answer and an incorrect answer, you task is to identify the first incorrect step from the incorrect answer. The "first incorrect step" means all reasoning up to that point is accurate, but the error begins at this specific step.\\

Question: \texttt{<QUESTION>}\\
Reference Answer: \texttt{<REFERENCE>}\\
Incorrect Answer: \texttt{<INCORRECT>}\\

After performing these tasks, You will output a json object containing the following fields:

\begin{lstlisting}[style=jsonstyle]
{
  "Justification": "string", // A brief justification for your output, 
  up to 100 words. 
  You need to explain 
  (1) why the identified first incorrect step is incorrect; 
  (2) why the reasoning up to this specific step is correct and 
  (3) how the corrected step resolves the issue, aligning with the reference answer, 
  maintaining the logical flow and progressing to the final answer.

  "First incorrect step": "string", // The explanation in the incorrect answer consists of multiple reasoning steps. Please identify the first incorrect reasoning step. It should be a piece of text directly and exactly quoted from the incorrect answer. It should be an intermediate step rather than the final answer

  "Reasoning up to incorrect": "string", // From the incorrect answer, give the correct reasoning steps up to the first incorrect step. This should be directly and exactly quoted from the incorrect answer.

  "Step correction": "string", //  Replace the identified incorrect step with a single, clear, and correct step. This step should directly address and correct the error, explicitly providing the correct reasoning without requring for more information or challenging the question. It should effectively answer the question, "What is the next reasoning step?" given on the question and the identied "Reasoning up tp incorrect". It should help progress to the final answer.

}
\end{lstlisting}

\end{tcolorbox}
\caption{Prompt for SCP}
\label{fig.prompt_for_genprm}
\end{figure*}

\end{document}